%% file: main.tex
\documentclass[10pt,journal,compsoc]{IEEEtran}

\ifCLASSOPTIONcompsoc
  \usepackage[nocompress]{cite}
\else
  \usepackage{cite}
\fi

\input{commands}
\usepackage[table]{xcolor}
\usepackage{graphicx}

\usepackage{standalone}
\usepackage{amsmath,amssymb,amsfonts}
\usepackage{algorithmic}
\usepackage{textcomp}
\usepackage{babel}
\usepackage{setspace}
\usepackage{balance}

\usepackage{varwidth}
\usepackage{verbatim}

\usepackage{tikz}
\usetikzlibrary{positioning,shapes.geometric,fit,calc}
\usepackage{pgfplots}
\pgfplotsset{compat=1.16, every non boxed x axis/.append style={x axis line style=-},
     every non boxed y axis/.append style={y axis line style=-}}
     
\usepackage[edges]{forest}
\usepackage{caption}

\tikzset{parent/.style={align=center,text width=3cm,rounded corners=3pt},
    child/.style={align=center,text width=3cm,rounded corners=3pt}
}

\colorlet{col1}{white}
\colorlet{col2}{gray!15}
\colorlet{col3}{gray!30}
\colorlet{col4}{gray!40}
\colorlet{col5}{gray!50}

\colorlet{transparent}{white!0}
\definecolor{ieeeblue}{RGB}{0,120,177}
\definecolor{accessblue}{cmyk}{1 0.3 0 0.2}
\definecolor{accessDarkBlue}{HTML}{006699}
\definecolor{tabletopheader}{gray}{.6}
\colorlet{tableheader}{ieeeblue}
\definecolor{tableoddrow}{gray}{.95}
\definecolor{tableevenrow}{gray}{.9}

\usepackage[colorlinks = true,
            linkcolor= ieeeblue,
            urlcolor  = ieeeblue,
            citecolor = ieeeblue,
            anchorcolor = ieeeblue]{hyperref}

\usepackage[framemethod=TikZ]{mdframed}   
\newmdenv[%
    linecolor=gray,leftmargin=20,%
    rightmargin=20,
    backgroundcolor=gray!30,%
    nobreak=true,%
    skipabove=20,
    skipbelow=20
]{greybox}

\newmdenv[%
    linecolor=accessDarkBlue,
    linewidth=2pt,
    innertopmargin=8,
    innerbottommargin=8,
    innerleftmargin=8,
    innerrightmargin=8,
    nobreak=true,%
    skipabove=20,
    skipbelow=20
]{darkbluebox}

\usepackage{multirow}
\usepackage{booktabs}
\arrayrulecolor{black!0} 
\newcolumntype{L}{>{\ifnumequal{\rownum}{1}{\sffamily\color{white}}}l}
\newcolumntype{C}{>{\ifnumequal{\rownum}{1}{\sffamily\color{white}}}c}
\newcolumntype{R}{>{\ifnumequal{\rownum}{1}{\sffamily\color{white}}}r}
\newcolumntype{Q}{>{\ifnumless{\rownum}{3}{\sffamily\color{white}}}l}
\newcolumntype{D}{>{\ifnumless{\rownum}{3}{\sffamily\color{white}}}c}
\newcolumntype{E}{>{\ifnumless{\rownum}{3}{\sffamily\color{white}}}r}

\def\BibTeX{{\rm B\kern-.05em{\sc i\kern-.025em b}\kern-.08em
    T\kern-.1667em\lower.7ex\hbox{E}\kern-.125emX}}
    
\ifCLASSOPTIONcaptionsoff
    \usepackage[nomarkers]{endfloat}
    \let\MYoriglatexcaption\caption
    \renewcommand{\caption}[2][\relax]{\MYoriglatexcaption[#2]{#2}}
\fi

\usepackage{url}

\begin{document}
%
\newcommand{\Title}{Decentral and Incentivized Federated Learning Frameworks: A Systematic Literature Review}
\title{\Title}
%
%
%

\author{Leon Witt,
        Mathis Heyer,
        Kentaroh Toyoda,~\IEEEmembership{Member,~IEEE,}
        Wojciech Samek$^*$,~\IEEEmembership{Member,~IEEE,} 
        Dan Li$^*$
\IEEEcompsocitemizethanks{
\IEEEcompsocthanksitem L. Witt is with the Department of Computer Science, Tsinghua University, Beijing, China, and with the Department of Artificial Intelligence, Fraunhofer Heinrich Hertz Institute, 10587 Berlin, Germany.\protect\\
E-mail: leonmaximilianwitt@gmail.com
\IEEEcompsocthanksitem M. Heyer is with Tsinghua University, Beijing, China, and RWTH Aachen University, Aachen, Germany.
\IEEEcompsocthanksitem K. Toyoda is with A*STAR, Singapore, and Keio University, Japan.
\IEEEcompsocthanksitem W. Samek is with the Department of Electrical Engineering and Computer Science, Technical University of Berlin, 10587 Berlin, with the Department of Artificial Intelligence, Fraunhofer Heinrich Hertz Institute, 10587 Berlin, Germany, and with BIFOLD -- Berlin Institute for the Foundations of Learning and Data, 10587 Berlin, Germany.\protect\\
E-mail: wojciech.samek@hhi.fraunhofer.de
\IEEEcompsocthanksitem D. Li is with the Department of Computer Science, Tsinghua University, China.
}

}

\IEEEtitleabstractindextext{%
\begin{abstract}
The advent of Federated Learning (FL) has sparked a new paradigm of parallel and confidential decentralized Machine Learning (ML) with the potential of utilizing the computational power of a vast number of Internet of Things (IoT), mobile, and edge devices without data leaving the respective device, thus ensuring privacy by design. Yet, simple Federated Learning Frameworks (FLF)  naively assume an honest central server and altruistic client participation. In order to scale this new paradigm beyond small groups of already entrusted entities towards mass adoption, FLFs must be (i) truly decentralized, and (ii) incentivized to participants. This systematic literature review is the first to analyze FLFs that holistically apply both, blockchain technology to decentralize the process and reward mechanisms to incentivize participation.
\rawresults publications were retrieved by querying 12 major scientific databases. After a systematic filtering process, \finalresults articles remained for an in-depth examination following our five research questions. To ensure the correctness of our findings, we verified the examination results with the respective authors. Although having the potential to direct the future of distributed and secure Artificial Intelligence, none of the analyzed FLFs is production-ready. The approaches vary heavily in terms of use cases, system design, solved issues, and thoroughness. We provide a systematic approach to classify and quantify differences between FLFs, expose limitations of current works and derive future directions for research in this novel domain. 


\end{abstract}

}

\maketitle

\IEEEdisplaynontitleabstractindextext

%
\IEEEpeerreviewmaketitle


%
%
%
%
\input{1_intro}
\input{2_preliminaries}

\input{3_related-surveys}
\input{4_results}
\input{5_future_research_directions}
\input{6_conclusion}

\appendices

\bibliographystyle{IEEEtran}
\bibliography{bibliography}

\begin{IEEEbiography}[{\includegraphics[width=1in,height=1.25in,clip,keepaspectratio]{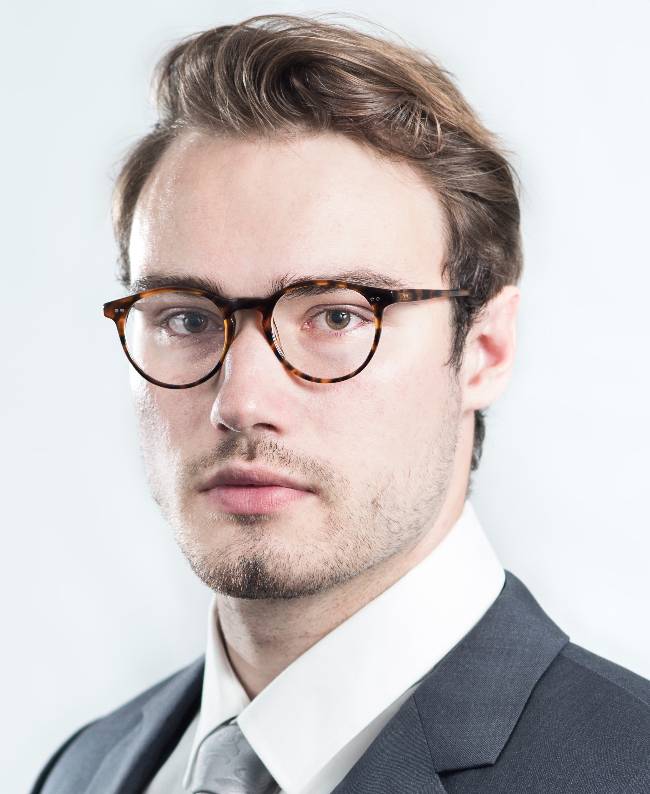}}]{Leon Witt} is a Ph.D. student at the Department of Computer Science and Technology at Tsinghua University in Beijing. He obtained a Master's degree in Mechanical Engineering and Business Administration from RWTH Aachen, Germany with exchange semesters in Zurich and Los Angeles. He obtained a second Master's degree in Industrial Engineering at Tsinghua University in 2017. His research interests lie at the intersection of federated artificial intelligence, blockchain, and mechanism design.
\end{IEEEbiography}

\begin{IEEEbiography}[{\includegraphics[width=1in,height=1.25in,clip,keepaspectratio]{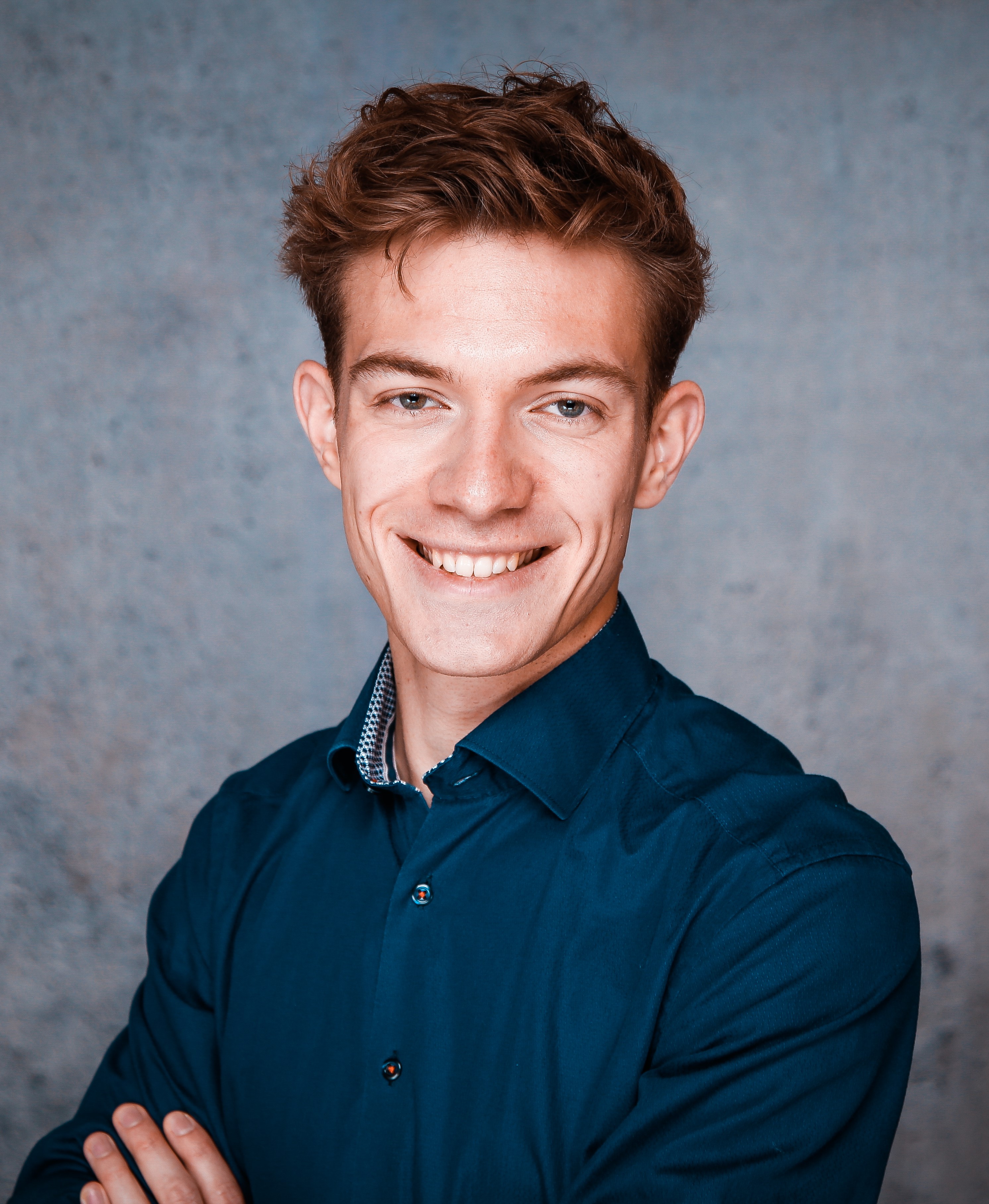}}]{Mathis Heyer}
is a Master's degree student in Chemical Engineering and Industrial Engineering at RWTH Aachen University, Germany, and Tsinghua University, China. He obtained his Bachelor's degree in Mechanical Engineering from RWTH Aachen University in 2021. As a visiting student, he spent the academic year 2019/2020 at Carnegie Mellon University, USA. His current research interests lie in the applications of artificial intelligence in fields such as chemical engineering and industrial engineering.
\end{IEEEbiography}

\begin{IEEEbiography}[{\includegraphics[width=1in,height=1.25in,clip,keepaspectratio]{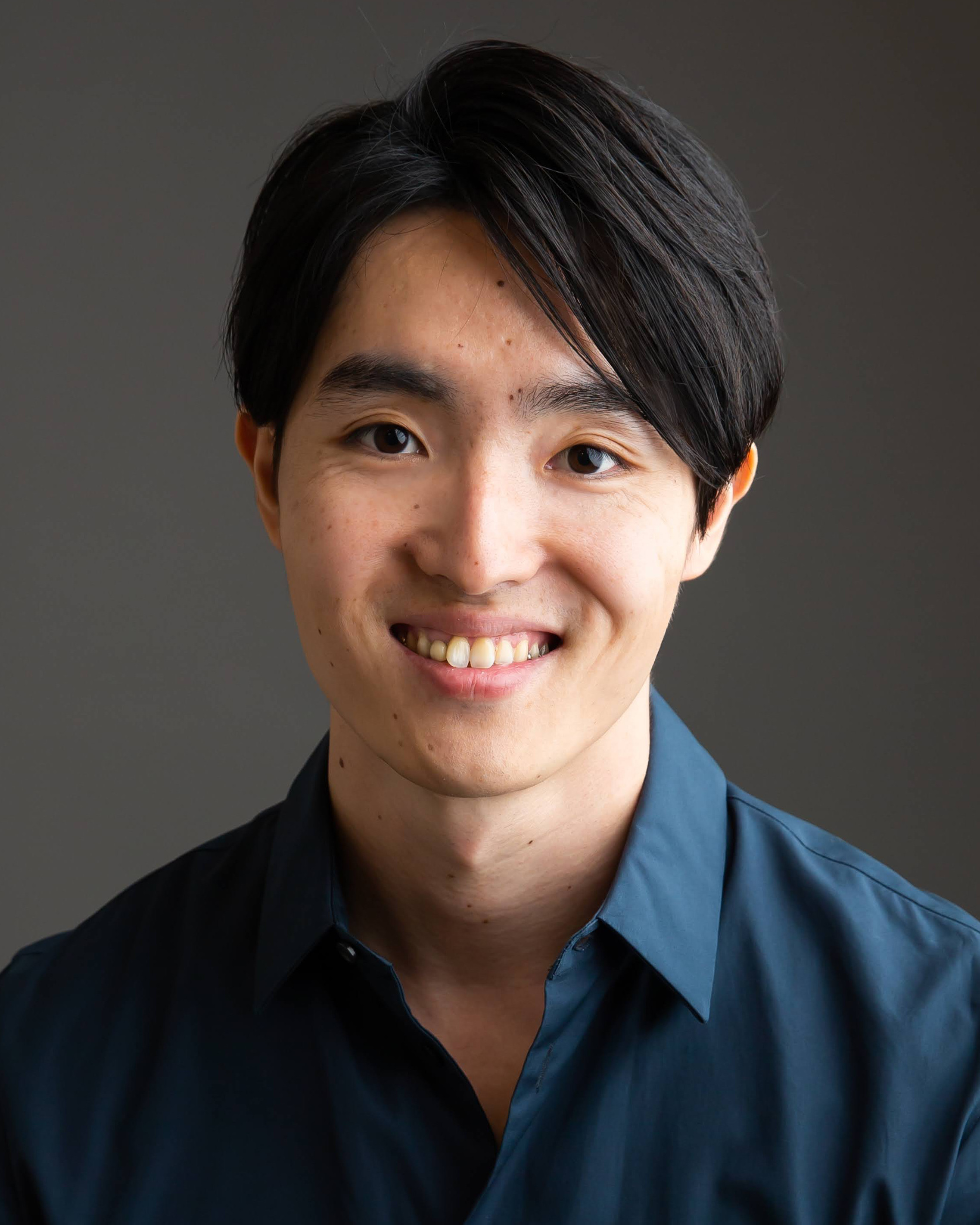}}]{Kentaroh Toyoda}
was born in Tokyo, Japan in 1988. He received B.E., M.E., and Ph.D. (Engineering) degrees in the Department of Information and Computer Science, Keio University, Yokohama, Japan, in 2011, 2013, and 2016, respectively. He was an assistant professor at Keio University from Apr. 2016 to Mar. 2019 and is currently a scientist at A*STAR, Singapore. His research interests include blockchain, mechanism design, security and privacy, and data analysis.
\end{IEEEbiography}


\begin{IEEEbiography}[{\includegraphics[width=1in,height=1.25in,clip,keepaspectratio]{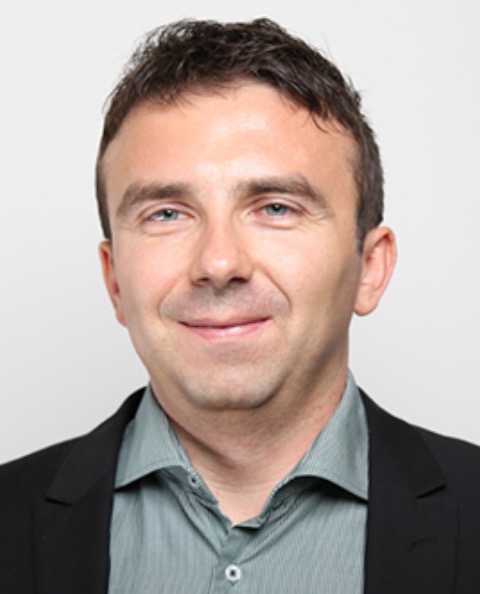}}]{Wojciech Samek}
(M'13) is a professor at the Department of Electrical Engineering and Computer Science at the Technical University of Berlin and is jointly heading the Department of Artificial Intelligence at Fraunhofer Heinrich Hertz Institute (HHI), Berlin, Germany. He received the Master's degree in computer science from the Humboldt University of Berlin, Germany, in 2010, and a Ph.D.\ degree from the Technical University of Berlin in 2014. He is  Associate Faculty with the Berlin Institute for the Foundation of Learning and Data (BIFOLD) and the ELLIS Unit Berlin. He has co-authored more than 150 peer-reviewed publications, several of which were listed by Thomson Reuters as highly cited papers. He is Senior Editor at IEEE TNNLS, and serves on the editorial boards of Pattern Recognition. Furthermore, he is an elected member of the IEEE MLSP Technical Committee and a recipient of multiple best paper awards, including the 2020 Pattern Recognition Best Paper Award and the 2022 Digital Signal Processing Best Paper Prize. His research interest includes deep learning, explainable and thrustworthy AI, and federated learning.
\end{IEEEbiography}

\begin{IEEEbiography}[{\includegraphics[width=1in,height=1.25in,clip,keepaspectratio]{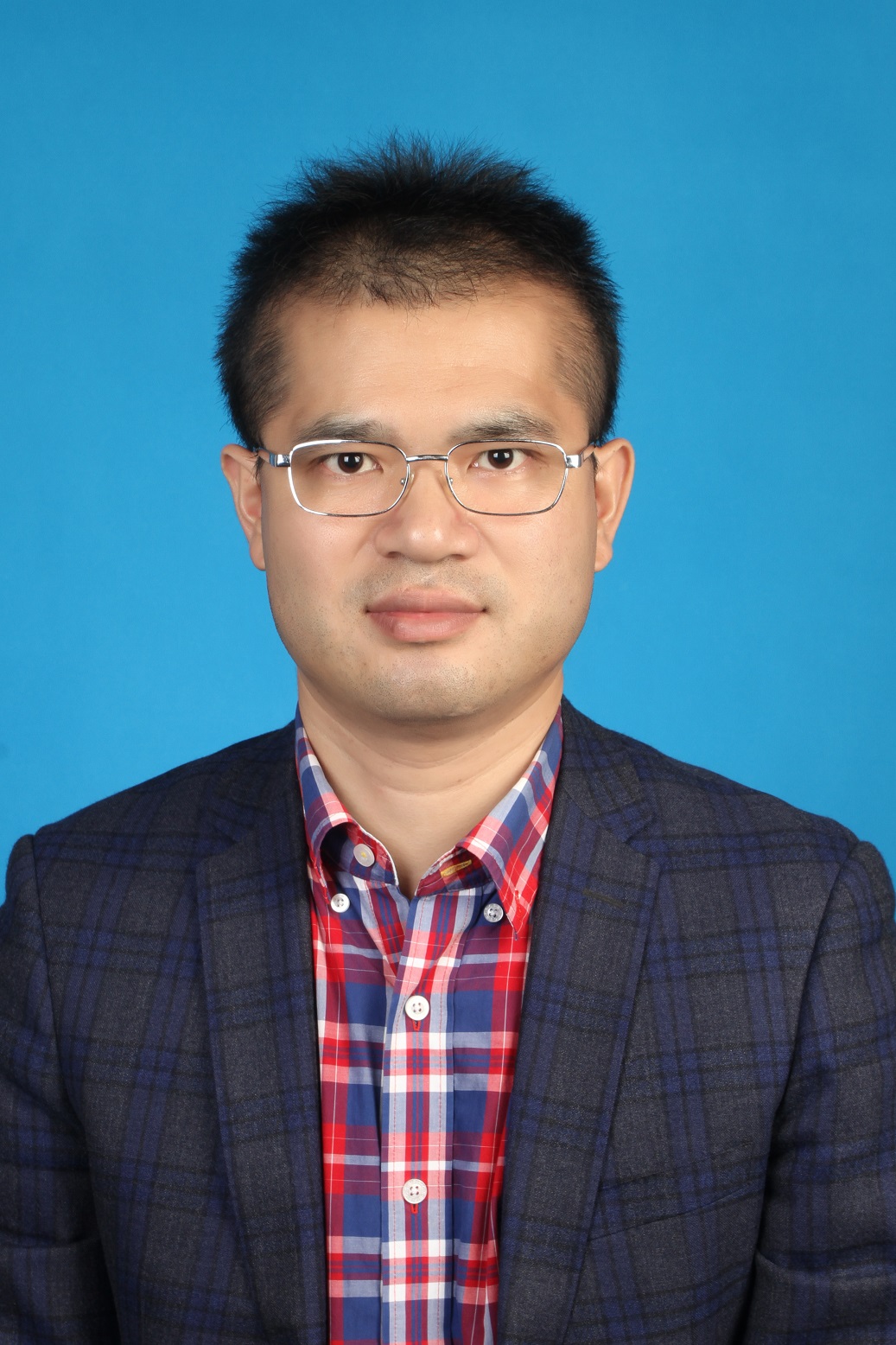}}]{Dan Li}
is currently a full professor at the Department of Computer Science and Technology at Tsinghua University-authored the NASP (Network Architecture, System and Protocols) research group, which is part of the networking research lab. He joined the faculty of Tsinghua University in March 2010, after two years working in the Wireless \& Networking Group of Microsoft Research Asia as an associate researcher. His main research direction includes trustworthy internet, data center networks and data-driven networking.
\end{IEEEbiography}




\end{document}


%
\newcommand{\Title}{Decentral and Incentivized Federated Learning Frameworks: A Systematic Literature Review\\[+6px]{\sc Supplementary Materials}}
\title{\Title}
%
%
%

\author{Leon Witt,
        Mathis Heyer,
        Kentaroh Toyoda,~\IEEEmembership{Member,~IEEE,}
        Wojciech Samek$^*$,~\IEEEmembership{Member,~IEEE,} and 
        Dan Li$^*$
\IEEEcompsocitemizethanks{
\IEEEcompsocthanksitem L. Witt is with the Department of Computer Science, Tsinghua University, Beijing, China, and with the Department of Artificial Intelligence, Fraunhofer Heinrich Hertz Institute, 10587 Berlin, Germany.\protect\\
E-mail: leonmaximilianwitt@gmail.com
\IEEEcompsocthanksitem M. Heyer is with Tsinghua University, Beijing, China, and RWTH Aachen University, Aachen, Germany.
\IEEEcompsocthanksitem K. Toyoda is with A*STAR, Singapore, and Keio University, Japan.
\IEEEcompsocthanksitem W. Samek is with the Department of Electrical Engineering and Computer Science, Technical University of Berlin, 10587 Berlin, with the Department of Artificial Intelligence, Fraunhofer Heinrich Hertz Institute, 10587 Berlin, Germany, and with BIFOLD -- Berlin Institute for the Foundations of Learning and Data, 10587 Berlin, Germany.\protect\\
E-mail: wojciech.samek@hhi.fraunhofer.de
\IEEEcompsocthanksitem D. Li is with the Department of Computer Science, Tsinghua University, China.
}
}

%
%

\markboth{IEEE Transactions on Parallel and Distributed Systems -- Supplementary Materials}%
{Witt \MakeLowercase{\textit{et al.}}: \Title}
%



\maketitle

\section{Related Surveys and Motivation of This Paper}
\label{sec:related-work}
We have identified several survey papers in the context of either mechanism design and FL \cite{SP_FL_MD,SP_MD_IEEE,SP_MD_ARXIV} or blockchain and FL \cite{SP_Blockchain_IEEE, SP_MD_ARXIV, SP_FL_TECHRXIV}. \tablename~\ref{table:comparison_of_related_survey_papers} shows the comparison of the related survey papers and our own.


Hou \etal surveyed the state-of-the-art blockchain-enabled FL methods \cite{SP_Blockchain_IEEE}. They focused on how blockchain technologies are leveraged for FL and summarized them based on the types of blockchain (public or private), consensus algorithms, solved issues and target applications.

The other related survey papers focus on incentive mechanisms for FL \cite{SP_MD_IEEE, SP_MD_ARXIV, SP_FL_TECHRXIV, SP_FL_MD}.
Zhan \etal survey the incentive mechanism design dedicated to FL \cite{SP_MD_IEEE}. They summarize the state-of-the-art research efforts on the measures of clients' contribution, reputation and resource allocation in FL. Zeng \etal also survey the incentive mechanism design for FL \cite{SP_MD_ARXIV}. However, the difference is that they focus on incentive mechanisms such as Shapley values, Stackelberg game, auction, context theory and reinforcement learning. 
Ali \etal survey incentive mechanisms for FL \cite{SP_FL_TECHRXIV}. In addition to \cite{SP_MD_IEEE} and \cite{SP_MD_ARXIV}, they summarize involved actors (e.g. number of publishers and workers), evaluation datasets as well as advantages and disadvantages of the mechanisms and security considerations. Tu \etal \cite{SP_FL_MD} provide a comprehensive review for economic and game theoretic approaches to incentivize data owners to participate in FL. In particular, they cluster applications of Stackelberg games, non-cooperative games,  sealed-bid auction models, reverse action models as well as contract and matching theory for incentive MD in FL.
Nguyen \etal investigate opportunities and challenges of blockchain-based federated learning in edge computing \cite{FLBlockchainOpportunitiesChallanges}.


As revealed from our analysis and summarized in \tablename~\ref{table:comparison_of_related_survey_papers}, the existing survey papers lack a holistic view of decentralized \textit{and} incentivized federated learning which is crucial to spreading the new generation of widely adopted fair and trustworthy FL to the benefit of the data owner. 
To the best of our knowledge, this paper is the first systematic literature review on the topic of blockchain-enabled decentralized FL with incentive mechanisms.

\begin{table}[t]
    \centering
    \caption{Comparison of related survey papers.}
    \resizebox{\columnwidth}{!}{
    \label{table:comparison_of_related_survey_papers}
    \rowcolors{2}{tableoddrow}{tableevenrow}
    \begin{tabular}{C|CCCC}
        \rowcolor{tableheader} Ref. & Blockchain & Federated Learning & Incentive Mechanism & Experiment Analysis\\
        \hline
        \cite{SP_Blockchain_IEEE} & \checkmark & \checkmark & \xmark &\xmark\\
        \cite{SP_MD_IEEE} & Partially & \checkmark &\xmark &\xmark\\
        \cite{SP_MD_ARXIV} & Partially & \checkmark & \checkmark &\xmark\\
        \cite{SP_FL_TECHRXIV} &\xmark & \checkmark & \checkmark &  \xmark\\
        \cite{SP_FL_MD} &\xmark & \checkmark & \checkmark & \xmark\\
        \cite{FLBlockchainOpportunitiesChallanges} & Partially & \checkmark & Partially & \xmark \\
        \textbf{This work} & \checkmark (\textbf{Detailed}) & \checkmark & \checkmark & \checkmark\\
    \end{tabular}
    }
\end{table}

\section{Number of Papers by Publishers}
An overview of the number of papers found and included by publishers can be found in Table \ref{tab:summary_of_search_results}.
\begin{table}[t]
    \centering
    \caption{Number of papers found and included by publishers.}
    \label{tab:summary_of_search_results}
    \rowcolors{2}{tableoddrow}{tableevenrow}
    \resizebox{\columnwidth}{!}{
    \begin{tabular}{C|RR}
         \rowcolor{tableheader}
         Publisher & \#Papers Found & \#Papers Included \\
         \hline
         Springer & 167 & 5\\
         ScienceDirect & 95 & 1\\
         ACM & 30 & 4 \\
         IEEE & 29 & 21+2\footnote{2 Survey Papers} \\
         Emerald insight & 29 & 0 \\
         MDPI & 27 & 3 \\
         Hindawi & 20 & 2 \\
         Tayler \& Francis & 18 & 1\\
         SAGE journals & 5 & 0\\
         Inderscience & 2 & 0\\
         Wiley & 0 & 0 \\
         \hline
         Total & 422 & 39 \\
    \end{tabular}
    }
\end{table}

\section{Definition of Columns in the Overview Tables}
Table \ref{tab:columns_definition} defines the categories used in the explanatory tables 1-4 of the main paper.
\begin{table*}[t]
    \centering
    \caption{Definition of columns in the overview tables.}
    \label{tab:columns_definition}
    \rowcolors{2}{tableoddrow}{tableevenrow}
    \resizebox{\textwidth}{!}{
    \begin{tabular}{C|LLL}
         \rowcolor{tableheader}
         Table & Column & Definition & Possible Values and Examples \\
         \hline
         \cellcolor{tableoddrow} & Application & The field of application of a FLF & Generic, Internet of Things (IoT), \dots \\
         \cellcolor{tableoddrow} & Domains of key contributions & To which domains each paper contribute & System architecture, blockchain, \dots \\
         \cellcolor{tableoddrow} & FL setting & Whether a type of FL is given & CS (cross-silo), CD (cross-device)\\
         \cellcolor{tableoddrow} & Actors & The key actors in the FLF & Workers, aggregation servers, task publishers, \dots \\
         \multirow{-5}{*}{\cellcolor{tableoddrow} FL (\tablename~\ref{table:FL})} & Setup period & Whether information about the setup of a FLF is included (e.g. who deploys a blockchain) & \checkmark / \xmark\\
         \hline
         \cellcolor{tableevenrow} & Operation on BC & Operations taking place on-chain & Aggregation, payment, coordination, storage\\
         \cellcolor{tableevenrow} & BC framework & The underlying Blockchain framework applied within the FLF & Agnostic, novel, Ethereum, Corda \dots \\
         \cellcolor{tableevenrow} & Consensus mechanism & The consensus mechanism of BC applied in the FLF & PoW, PoS, PBFT, novel, \dots \\
         \cellcolor{tableevenrow} & Storage on BC & Information that is stored on the Blockchain. & Gradients/model parameters, reputation scores \dots \\
         \cellcolor{tableevenrow} & Storage quantification & Whether the amount of information stored on the BC is analyzed & \checkmark / \xmark \\
         \cellcolor{tableevenrow} & Storage off-chain & Whether to store data off-chain, such as Interplanetary File System (IPFS) & \checkmark / \xmark \\
         \multirow{-7}{*}{\cellcolor{tableevenrow}BC (\tablename~\ref{table:blockchain})} & Scalability & Whether scalability is considered. & \checkmark / \xmark \\
         \hline
         \cellcolor{tableoddrow} & Simulation & Whether profits, rewards or utilities were evaluated via simulation & \checkmark / \xmark \\
         \cellcolor{tableoddrow} & Theoretical analysis & Whether profits, rewards or utilities were theoretically analyzed & \checkmark(Used theory) / \xmark \\
         \cellcolor{tableoddrow} & Costs assumed in utility analysis & Cost elements considered for theoretical analysis & Energy consumption, \dots, or \xmark\ if not used \\
         \cellcolor{tableoddrow} & Metrics for contribution measurement & Metrics used to validate workers' contribution & Accuracy, data size, \dots or \xmark\ if not used \\
         \multirow{-5}{*}{\cellcolor{tableoddrow}IM (\tablename~\ref{table:incentive_mechanism})} & Validator & Entities that validate workers' contribution & Task requesters, smart contracts, \dots \\
         \hline
         \cellcolor{tableevenrow} & ML task & Machine Learning task of the experiment & Classification, regression,  \dots \\
         \cellcolor{tableevenrow} & Dataset & Datasets on which the experiments are conducted  & MNIST, CIFAR10,  \dots \\
         \cellcolor{tableevenrow} & Number of Clients & Number of clients within the experiments & \\
         \cellcolor{tableevenrow} & FL algorithm & FL algorithm applied within the experiments & \texttt{FedAvg},\texttt{FedProx}  \dots \\
         \cellcolor{tableevenrow} & Privacy protection & Whether additional privacy protection methods are applied within the experiments & \checkmark / \xmark  \\
         \cellcolor{tableevenrow} & Non-IID data & Whether experiments are conducted under the non-IID condition & \checkmark / \xmark \\
         \cellcolor{tableevenrow}& Adversaries & Whether the FNFs robustness is tested with malicious agents &  \checkmark / \xmark \\
         \cellcolor{tableevenrow} & BC implementation & Whether the blockchain part of the FNF is implemented for conducting experiments & \checkmark / \xmark \\
         \multirow{-9}{*}{\cellcolor{tableevenrow}Experiments (\tablename~\ref{table:experiments})} & Performance & Whether the performance within the experiments of the FL model is measured & \checkmark / \xmark   \\
         \hline
    \end{tabular}
    }
\end{table*}

\section{Privacy-preserving concepts employed in survey papers}
Figure \ref{abb:PrivacyClassification} gives an overview of the privacy-preserving concepts employed in survey papers.
\begin{figure}
    \centering
    \resizebox*{0.95\linewidth}{!}{%
        \begin{forest}
            forked edges,
            for tree={
                grow'=east,
                draw,
                rounded corners,
                text width=2cm,
                node options={align=center},
            }       
            [Privacy-preserving \\concepts in the \\surveyed papers, fill=col5, parent, s sep=1cm
            [Update \\masking \\\cite{LW10_refiner, MH11_Desai2021, Bao2019FLChain, LW6_Rahmadika2020}, fill=col4
            [Multi-Party \\Computation (MPC), for tree={child, fill=col3}
            [Google's Secure \\Aggregation (SA) \\\cite{MH4_Ma2021, LW9_FedCoin}, for tree={child, fill=col2}]
            [Yao's garbeled \\ circuits, fill=col2]
            [Homomorphic \\Encryption (HE)\\\cite{MH9_He2021, MH14_Rathore2019, LW3_Kumar2020}, for tree={child, fill=col2}
            [Paillier \\\cite{MH12_Zhu2021, LW2_Weng2021}, fill=col1]
            [RSA, fill=col1]
            [DGHV\\\cite{LW8_Privacy_IoV}, fill=col1]
            [Elgamal \\\cite{MH13_Li2020},fill=col1]
            ]
            ]
            [Differential \\Privacy (DP) \\\cite{MH7_Mugunthan2022, LW5_FL_HomeAppliances_IoT, LW3_Kumar2020}, fill=col3]
            ]
            [Identity \\protection, fill=col4
            [Combination of HE and/or other cryptographic techniques, fill=col3
            [\cite{LW8_Privacy_IoV}: Zero-Knowledge \\Proof (ZKP) , for tree={child, fill=col2}]
            [\cite{LW12_Rahmadika2021_unlinkable}: Ring signature~\& RSA~\& Rabin~\& ECC , for tree={child, fill=col2}]
            [\cite{LW14_5G_Rahmadika2021}: Pairing-based cryptography~\& ECC , for tree={child, fill=col2}]
            [\cite{MH1_Chai2021}: Asymmetric cryptography~\& digital signatures , for tree={child, fill=col2}]
            ]
            ]
            ]
        \end{forest}
    }
    \caption{Privacy-preserving concepts employed in survey papers. \textit{\footnotesize (DGHV = Dijk-Gentry-Halevi-Vaikutanathan Algorithm, ECC = Elliptic Curve Cryptography, RSA = Rivest-Shamir-Adleman Cryptosystem. Papers with unspecified methods are added to next highest node.)}}
    \label{abb:PrivacyClassification}
\end{figure}

\bibliographystyle{IEEEtran}
\bibliography{bibliography}

%% file: commands.tex
\usepackage{pifont}

\usepackage{xspace}
\newcommand{\etal}{\textit{et~al.}~}

\newcommand{\one}{({\textit{i}})~}
\newcommand{\two}{({\textit{ii}})~}
\newcommand{\three}{({\textit{iii}})~}
\newcommand{\four}{({\textit{iv}})~}
\newcommand{\five}{({\textit{v}})~}
\newcommand{\six}{({\textit{vi}})~}

\newcommand{\rawresults}{422~}
\newcommand{\finalresults}{40~}

\makeatletter
\renewcommand{\paragraph}[1]{\vspace*{0.03in}\noindent{\bf #1.}\hspace{0.25ex \@plus1ex \@minus.2ex}}
\makeatother

%% file: 1_intro.tex
\IEEEraisesectionheading{\section{Introduction}\label{sec:introduction}}


\IEEEPARstart{C}{entralized} platforms in the domains of search engines, mobile applications, social media, chat, music and retail have been dominating the respective industries over the past decades. Business models where digital services are exchanged for user data have developed into high-revenue industries with a few single entities controlling the global market within the respective domains~\cite{GAFAM}. The resulting concentration of user data in a small number of entities, however, poses problems such as the risk of private data leaks~\cite{Wheatley.2016} or an increasing power imbalance in favor of market-dominating parties~\cite{GAFAMpower,GAFAM,Santesteban.2020} which has caused policymakers to enhance data protection for individuals~\cite{EUdataregulations2018}. The need for confidential AI extends beyond B2C markets, such as when entities within the health sector or Internet of Things (IoT) companies are not allowed to collaborate on a common AI model due to sensitive data.

A promising solution that enables the training of Machine Learning (ML) models with improved data security is Federated Learning (FL). In FL, complex models such as Deep Neural Networks (DNNs) are trained in a parallel and distributed fashion on multiple end devices with the training data remaining local at all times. Federated Averaging (\texttt{FedAvg})~\cite{BrendanMcMahan2017} is a widely applied algorithm for FL where a central authority aggregates a global model from the locally trained models in an iterative process. In theory, FL not only makes previously withheld sensitive data accessible to the machine learning process but also enables efficient training by taking advantage of the ever-increasing computational power of IoT and mobile devices.  However, the majority of FL research focuses on advancing the efficiency of the technology, yet incentives and decentralization are necessary requirements for many real-world FL applications, and the prerequisite for FL to evolve from academic research to real-world products with the potential to disrupt the vigorous data and AI industry \cite{AdvancesAndOpenProblemsInFederatedLearning}. In particular, incentives and decentralization address the two major design problems in FL: (i) the star topology of FL that introduces the risk for a single point of failure as well as for authority abuse and prohibits use-cases where equal power among participants is a mandatory requirement, and (ii) the lack of a practical reward system for contributions of participants that hinders this technology from scaling beyond small groups of already entrusted entities towards mass adoption.



Although many proposals of Incentivized and Decentralized Federated Learning Frameworks (FLFs)  exist, we have not yet seen any full-fledged production-level FLF. To enhance the development towards production readiness, we compared state-of-the-art solutions despite their heterogeneity in terms of assumptions, use cases, design choices, special focus, and thoroughness by providing a general and holistic comparison framework.

\begin{figure*}[t]
    \centering
    \includegraphics[width=\textwidth]{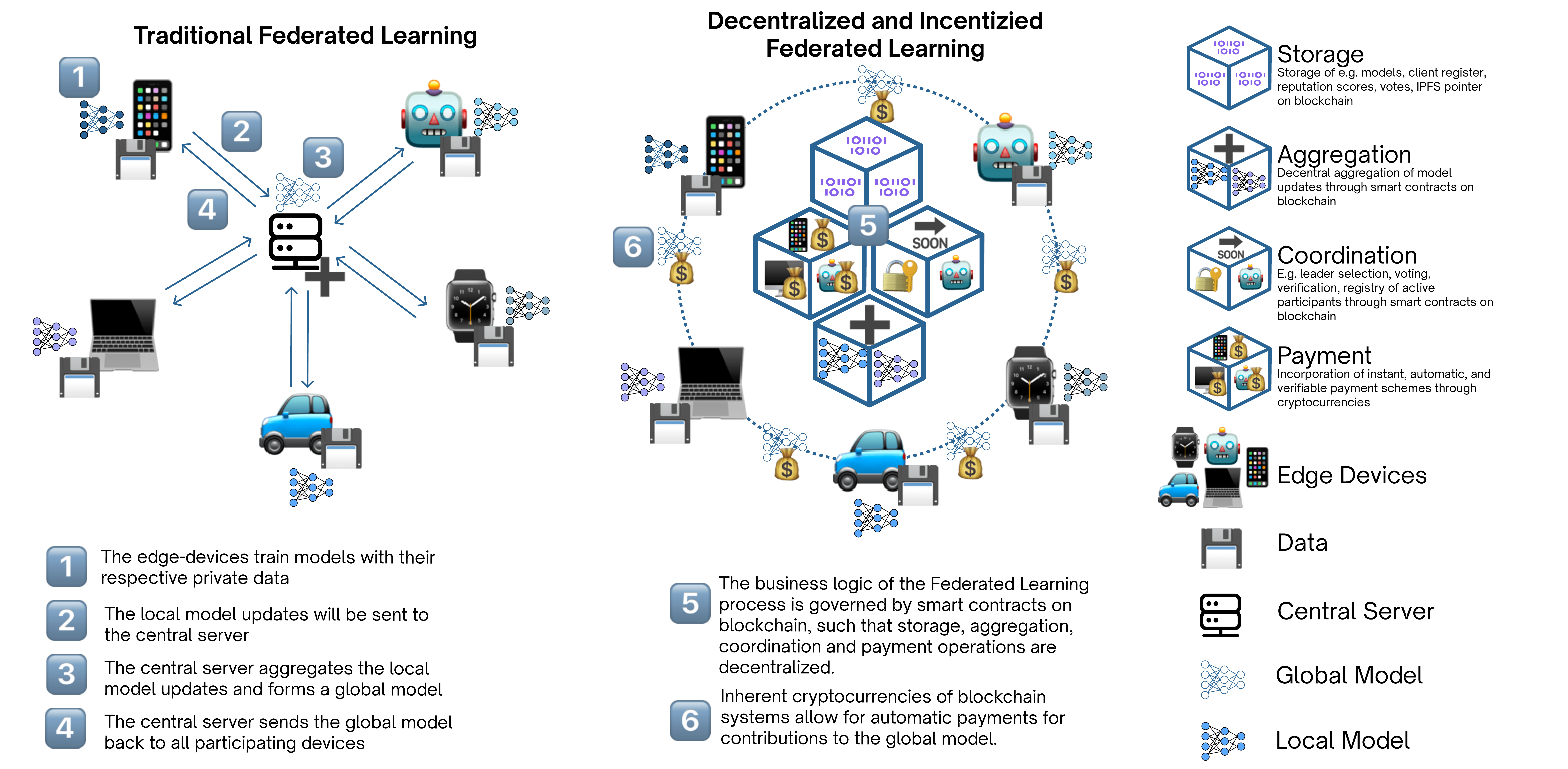}
    \caption{FL vs. Decentralized and Incentivized FLF.}
    \label{fig:FLvsFLFs}
\end{figure*}

Specifically, we undertake a Systematic Literature Review (SLR) examining all relevant articles from twelve scientific databases in the domain of computer science. \rawresults publications were queried from these databases and filtered for relevant contributions, resulting in \finalresults papers remaining after three filtering steps. To the best of our knowledge, this is the first comprehensive survey on the design of both decentralized and incentivized federated artificial intelligence systems. The contribution of this paper is threefold:

\begin{enumerate}
    \item The first comprehensive systematic survey study on the combined topic of decentralized and incentivized FL based on the standardized Preferred Reporting Items for Systematic Reviews and Meta-Analyses (PRISMA) process, ensuring transparency and reproducibility of the work.
    \item A novel comparison framework for an in-depth analysis of incentivized and decentralized FL frameworks which goes beyond existing survey papers by \one pointing out the limitations and assumptions of the chosen game-theoretic approaches, \two analyzing the existing solutions based on computational and storage overhead on the BC, and \three an in-depth analysis of the performed experiments.
    \item Based on this comparison, we have identified limitations of recent work, derived trade-offs in the design choices, and derived future research directions of decentralized and incentivized FLFs. 
\end{enumerate}

The remainder of this paper is structured as follows.
In Section~\ref{sec:background}, we present the technical background of distributed ledger technology and mechanism design in FL systems.
In Section~\ref{sec:related-work}, we provide an overview of existing surveys on this topic and their respective problem statements.
Section~\ref{sec:results} summarizes the findings of the SLR and answers research questions concerning general applications of the FLFs, BC features, Incentive Mechanisms (IMs), and experiments. We derive limitations and further research directions in Section~\ref{sec:future_research_directions}.
Finally, Section~\ref{sec:conclusion} concludes this literature review.

In the Appendix, we outline the details of the methodology, the search process, the selection process, the search terms, and the number of papers retrieved from the respective databases to ensure reproducibility.

%% file: 2_preliminaries.tex
\section{Preliminaries}
\label{sec:background}

The following sections discuss the fundamentals of FL, distributed ledger technology, and incentive design, and outline how these technologies are integrated in FLFs. 

\subsection{Federated Learning}\label{Federated Learning}
FL is a machine learning technique where multiple actors collaboratively train a joint machine learning model locally and in parallel, such that the individual training data does not leave the device as depicted on the left side of Figure~\ref{fig:FLvsFLFs}. This decentralized approach to machine learning was first introduced by Google in 2016~\cite{BrendanMcMahan2017} and addresses two key issues of machine learning: \one The high computational effort for model training is relocated from a single central actor to a network of data-owning training devices. \two As the training data remains on the edge devices, previously inaccessible data of privacy-concerned actors can be integrated into the training process. Thus, ``data islands'' are prevented.

FL tasks can be classified in horizontal and vertical settings. In horizontal FL, the actors possess the same type of information on different entities, whereas, in vertical FL, the actors have available different information features on the same entity. The latter comes with the additional challenge of data alignment and information exchange~\cite{SP_MD_ARXIV}. Furthermore, the FL setting can range from a few collaborating entities, i.e., Cross-Silo (CS), to a federated system of millions of devices, i.e., Cross-Device (CD). 

The Federated Averaging (\texttt{FedAvg}) algorithm~\cite{BrendanMcMahan2017} is a widely adopted optimization algorithm for FL. Its objective is to minimize the empirical risk of the global model $\theta$, such that

\begin{equation}
\arg\min_{\boldsymbol{\theta}} \sum_{i} \frac{|S_i|}{|S|} f_{i}(\boldsymbol{\theta})
\label{Eq:Optimization}
\end{equation}
where for each agent $i$, $f_{i}$ represents the loss function, $S_i$ is the set of indexes of data points on each client, and $S:=\bigcup_{i} S_i$ is the combined set of indexes of data points of all participants. For that, the calculated gradients from the respective local model training get aggregated in three communication rounds:
\begin{enumerate}
    \item A central server broadcasts a first model initialization $\theta_{init}$ to a subset of participating clients\footnote{Clients, workers, and agents are used interchangeably.}.
    \item These clients individually perform iterations of stochastic gradient descent over their local data to improve their respective local models $\theta_{i}$.
    \item In order to create a global model $\theta$, all individual models $\theta_{i}$ are then sent back to the server, where they are aggregated (e.g., by an averaging operation). This global model is used as the initialization point for the next communication round.
\end{enumerate}

Optimization algorithms for the FL case are open research \cite{AdvancesAndOpenProblemsInFederatedLearning} and variations of \texttt{FedAvg} exist, e.g., \texttt{FedBoost}~\cite{Fedboost}, \texttt{FedProx}~\cite{FedProx}, \texttt{FedNova}~\cite{FedNova}, \texttt{FedSTC}~\cite{SatTNNLS20} and \texttt{FetchSGD}~\cite{fetchSGD}.

\subsection{Blockchain: A Distributed Ledger Technology}

 A distributed ledger kept by nodes in a peer-to-peer network is referred to as blockchain, first invented by Satoshi Nakamoto through Bitcoin in 2008~\cite{nakamoto2008bitcoin}. Cryptographic connections of information enable resistance to alteration and immutability. A peer-to-peer consensus mechanism governs the network, obviating the requirement for central coordination~\cite{8693657}. The introduction of general-purpose BCs with smart contract capability supports Turing-completeness~\cite{wood2014ethereum} and has allowed for the creation of decentralized, immutable, and transparent business logic on top of the BC. Here, smart contracts are computer programs that are decentrally stored on a distributed BC network and automatically executed when a predetermined condition is met. 

 \subsubsection{BC data architecture}
Even though the data structure varies across different BC, the structure can roughly be categorized into six layers. 
 \begin{itemize}
     \item \textbf{Data Layer}:
     The data layer defines how new information will get stored and how the respective blocks are designed. Blocks typically contain a block header and block body \cite{wood2014ethereum, nakamoto2008bitcoin}. The block header is a collection of metadata about the block and a summary of the transactions included in the execution payload. In Ethereum, the block body is a bundled unit of information that include an ordered list of transactions and consensus-related information \cite{wood2014ethereum}.
     \item \textbf{Network Layer}: 
     BCs are peer-to-peer networks with nodes that require defined protocols for communication between them. The protocol comprises exchanging requests and answers between particular nodes (one-to-one communication) as well as "gossiping" information (one-to-many communication) through the network \cite{EthereumNetworkLayer}. To make sure they are delivering and receiving the right information, each node must abide by a set of networking regulations.
     \item \textbf{Consensus Layer}:
     The consensus mechanism refers to the entire stack of protocols, incentives, and ideas that allow a network of nodes to agree on the state of a BC. 
     The two major types of consensus mechanisms for public BCs are Proof of Work (PoW) and Proof of Stake (PoS). PoW is used in Bitcoin \cite{nakamoto2008bitcoin} and the former version of Ethereum \cite{wood2014ethereum}, where nodes in the network compete to find a specific hash value below a given number to prove that a certain amount of a specific computational effort has been expended in order to add blocks to the network. The fact that it would take 51\% of the network's computer power to commit fraud on the chain ensures the network's security \cite{nakamoto2008bitcoin}.
     In PoS-based consensus mechanisms, blocks can be added by randomly chosen validators who have staked significant amounts of cryptocurrencies. The system is designed such that the staked cryptocurrencies get slashed when the validators act maliciously, securing the network crypto-economically \cite{EthereumPos}.
     \item \textbf{Incentive Layer}:
     To incentivize participation in the consensus mechanism, BC systems contain an inherent cryptocurrency reward for either contributing computational effort (i.e., PoW) or staking cryptocurrencies (i.e., PoS). 
     FL-specific BC systems may reward operations beyond securing the BC such as the storage of ML models \cite{LW2_Weng2021, LW8_Privacy_IoV, LW11_RewardResponseGame}, the aggregation of gradients \cite{LW1_witt2021rewardbased, LW2_Weng2021, KT01_Toyoda2019BigData, KT02_Toyoda2020Access, KT02_Toyoda2020Access} or contribution calculations\cite{LW1_witt2021rewardbased,KT01_Toyoda2019BigData, LW9_FedCoin}.
     \item \textbf{Contract Layer}:
     Smart contracts are simple programs that run on the respective virtual machine of the BC. The Ethereum Virtual Machine (EVM) is a Turing-complete environment for smart contracts most commonly used across other BCs such as Polygon~\cite{MaticWhitepaper}, BNB Chain~\cite{BinanceChainWhitepaper} and Avalanche~\cite{AvalancheWhitepaper}.  
     A smart contract is a collection of code (its functions) and data (its state) that resides at a specific address on the respective BC.
     \item \textbf{Application Layer}:
     Decentralized Applications (Dapps) on top of smart contracts cannot be censored, allow for anonymous participation, have zero downtime, and are compatible with other Dapps on the same BC.
 \end{itemize}

 \subsubsection{BC-based FL to ensure equal power}
 Due to its intrinsic features, distributed ledger technology is capable of mitigating open issues in the FL context, namely:

\begin{itemize}
    \item \textbf{Decentralization}: Workers are subject to a power imbalance and a single point of failure (SPoF) in server-worker topologies. A malicious server might refuse to pay reward payments or exclude employees at will. Furthermore, a server-worker architecture is incompatible with a situation in which numerous entities have a shared and equal stake in the advancement of their respective models. BC technologies' decentralization provides a federal system for entities of equal authority without the need for a central server.  
    \item \textbf{Transparency and Immutability}: Data on the BC can only be updated, not erased, as every peer in the system shares the same data. In a FL environment, a clear and immutable reward system ensures worker trust. On the other side, each client is audited, and as a result, can be held accountable for malevolent activity.
    \item \textbf{Cryptocurrency}: Many general-purpose BC systems include cryptocurrency capabilities, such as the ability to incorporate payment schemes within the smart contract's business logic. Workers can be rewarded instantly, automatically, and deterministically based on the FL system's reward mechanism.
\end{itemize}

Therefore, BC systems \cite{wood2014ethereum,polkadotwhitepaper, HyperledgerFabric} have the potential to mitigate the first issue of FL by ensuring trust through their inherent properties of immutability and transparency. They enable decentralized federations to mitigate dependencies on a central authority. 

Figure~\ref{fig:FLvsFLFs} (right) depicts the four major functions BC helps to facilitate in the FL process, namely aggregation, coordination, storage, and payment. 
\begin{itemize}
    \item \textbf{Aggregation}: In regular \texttt{FedAvg}-based FL, a central authority collects and aggregates the clients' gradients. BC can decentralize the process by performing the aggregation on-chain. 
    \item \textbf{Coordination}: Leader selection for the aggregation process, voting, verification, and onboarding of new clients are necessary for real-world FL but undefined steps in contemporary research. BC can provide a trustless and transparent infrastructure for those steps. 
    \item \textbf{Storage}: BC provides immutable and transparent storage for information where access is shared among clients. 
    \item \textbf{Payment}: BCs inherent functionality of cryptocurrencies allows for automated payment schemes to reward clients for the exerted effort. These four major operations of BC in FL are discussed in detail in Section~\ref{sec:results_blockchain}.
\end{itemize}



\subsection{Incentive Mechanism}
\label{sec:mechanism_design}
FL use cases where pseudo-anonymous clients are expected to participate and invest their data and computational power cannot assume altruism but require compensation in any real-world scenario. 
 Mechanism Design (MD), which is a field of economics, attempts to implement a protocol, system, or rule so that the desired situation (e.g., every participant contributes informed truthful model updates) is realized in a strategic setting, assuming that each participant acts rationally in a game theoretic sense~\cite{nisan2007introduction}.

The purpose of incorporating MD into FL is to incentivize clients to \one put actual effort into obtaining real and high-quality signals (i.e., training the model on local data) and \two submit model updates truthfully despite not being monitored directly. Such incentives can be distributed using BC infrastructure and their underlying cryptocurrencies. An appropriately designed mechanism ensures a desired equilibrium when every worker acts rationally and in their own best interest. Moreover, a mechanism ideally has low complexity and is self-organizing, avoiding the need for Trusted Execution Environments (TEE)~\cite{TXX} or secure multi-party computation (MPC), yet makes assumptions about the degree of information available.

The process of designing a FL protocol with MD consists of \one designing a mechanism and \two a theoretical analysis. The former determines the whole procedure of FL including a reward policy. The reward policy defines \one how to measure clients' contribution to the overall model and \two how to distribute rewards. Measuring contributions in FL is challenging since the aggregated gradients do not reveal explicit information about their effect on the overall performance. Additionally, a myriad of design choices for reward distribution exists, e.g., whether rewards should be given to the top contributor (i.e., winner-takes-all) or to multiple workers where rewards are equally distributed among all contributors or unequally distributed based on the workers' contribution.


\subsubsection{Theories behind mechanism design}
We classify underlying theories for mechanism design broadly into two categories, namely \one game theory and \two auctions, based on~\cite{SP_FL_MD}. 

The theory assumes that the clients' utility $U$ is defined by expected profits $\Pi$ (e.g., prizes) minus costs $C$ (e.g., costs of model training and data collection).
 \begin{equation}
 \label{eq:utilityfunction}
    U =  \Pi -  C
\end{equation}
 
Assuming individual rationality, clients choose their actions to maximize their utilities. In this context, the interactions of choices that produce outcomes concerning utilities are referred to as games in the scientific literature. Games can be classified into cooperative and non-cooperative. A non-cooperative game is a game where each client individually determines their strategies so that their utilities are maximized, while a cooperative game maximizes the utility of the group. A game is called imperfect when a client does not know the others' information (e.g., utilities, strategies, etc.). When all the clients know others' information, such a game is called perfect. 
Popular game-theoretic methods applied in FL are Stackelberg games, contest theory, and contract theory. 

\textbf{Stackelberg game}: A leader (e.g., a task requester) determines their strategy, and followers (i.e., clients or workers) determine theirs according to the leader's action~\cite{stackelberg1952theory}. A task requester can be a leader who determines a reward first, and clients determine their effort and how much they should exert based on the condition. Seminal work on Stackelberg games includes, e.g., Khan~\etal who motivate the modeling of FL as Stackelberg game and propose an IM based on their best response algorithm~\cite{Khan2020IM}. Another example of a Stackelberg game-based incentive mechanism in centralized FL settings can be found with Zhang~\etal~\cite{Zhan2020DL}. The authors tackle the challenges of information secrecy and contribution measurement by training a deep reinforcement learning-based IM that allows optimal pricing for the central server and optimal training strategies for the participating clients.  

\textbf{Contest Theory}: Contrary to Stackelberg games, clients need to exert efforts before joining the contest \cite{Vojnovic2017contest}. The process of FL can be seen as a contest as clients must train a model with their own local datasets while they are not guaranteed to receive prizes. 

\textbf{Contract Theory}: In contract theory, an employer has to agree on a contract with employees given the situation that the employees may claim false capabilities~\cite{bolton2004contract}. This could be the case in FL as task requesters do not exactly know clients' capabilities (e.g., cost, computational resources, etc.).

\textbf{Auctions}: Auctions are applicable in designing the mechanisms of FLF as they optimally allocate resources such as computational resources or amount of data, based on clients' reports. A task requester posts a FL task, potential clients bid with sealed information such as computational cost and resource, and the requester assigns a FL task to winning clients. There are several auctions to determine winners (e.g., first-price sealed-bid (FPSB) auction, second-price sealed-bid (SPSB) auction, Vickrey-Clarke-Groves (VCG) auction, all-pay auctions)~\cite{krishna2009auction}. A detailed analysis of all auction types is beyond the scope of this paper which is limited to the most popular auctions applied in FLFs: a FPSB auction is an auction where no bidders know others' bids and the highest bidder pays the price that they bid. A SPSB auction is similar to the FPSB, but the highest bidder only needs to pay the price that the second highest bid. A VCG auction is a sealed-bid auction for multiple resources. It is designed to achieve socially optimal resource allocation by charging winners of an auction the social loss they cause to others. This prevents clients from bidding their false valuations to win. An all-pay auction is an auction where all bidders need to pay regardless of whether or not they win. A Tullock contest is one of the most famous all-pay auctions~\cite{tullock2001efficient}.

\subsubsection{Desirable properties}
Game theory and auctions provide a strong guarantee that a designed mechanism possesses desirable properties. Zeng~\etal summarized the main properties that a mechanism should possess in FL, namely incentive compatibility (IC), individual rationality (IR), Pareto efficiency (PE), collusion resistance (CR) and fairness, balanced budget (BB)~\cite{SP_MD_ARXIV}.
IC is fulfilled when entities cannot be better off by deviating from their optimal strategies, and IR refers to the assumption that contributors would not participate if their respective utility was negative as in Eq.~\eqref{eq:utilityfunction}. 
A game is PE when it guarantees that the sum of profits is maximized. CR is achieved when no participant can be better off by colluding with others.
A game is said to be fair when fairness, e.g., a payoff to the contribution, is preserved~\cite{Yu2020fairness}. Finally, a game is BB when it is sustainable without external money inflows. Section~\ref{sec:results_incentive_mechanism} discusses how these theories and properties are adopted in the FLFs' MD.


%% file: 3_related-surveys.tex
\section{Related Surveys}
To the best of our knowledge, this is the first analysis of holistic frameworks for fully decentralized FL with rewards for the participating clients. Yet, we have identified several survey papers in the context of either MD and FL~\cite{SP_FL_MD,SP_MD_IEEE,SP_MD_ARXIV} or BC and FL~\cite{SP_Blockchain_IEEE, SP_MD_ARXIV, SP_FL_TECHRXIV}. \tablename~\ref{table:comparison_of_related_survey_papers} shows the comparison of the related survey papers and our own.

\label{sec:related-work}
\begin{table}[t]
    \centering
    \caption{Comparison of related survey papers.}
    \resizebox{\columnwidth}{!}{
    \label{table:comparison_of_related_survey_papers}
    \rowcolors{2}{tableoddrow}{tableevenrow}
    \begin{tabular}{C|CCCC}
        \rowcolor{tableheader} Ref. & BC & FL & IM & Experiment Analysis\\
        \hline
        \cite{SP_Blockchain_IEEE} & \checkmark & \checkmark &  &\\
        \cite{SP_MD_IEEE} & Partially & \checkmark & &\\
        \cite{SP_MD_ARXIV} & Partially & \checkmark & \checkmark &\\
        \cite{SP_FL_TECHRXIV} & & \checkmark & \checkmark &  \\
        \cite{SP_FL_MD} & & \checkmark & \checkmark & \\
        \cite{FLBlockchainOpportunitiesChallanges} & Partially & \checkmark & Partially &  \\
        \cite{BlockchainFLSurvey} & \checkmark & \checkmark & Partially &  \\
        \textbf{This work} & \checkmark (\textbf{Detailed}) & \checkmark & \checkmark & \checkmark\\
    \end{tabular}
    }
\end{table}


Hou~\etal investigate the state-of-the-art BC-enabled FL methods~\cite{SP_Blockchain_IEEE}. They focus on how BC technologies are leveraged for FL and summarized them based on the types of BC (public or private), consensus algorithms, solved issues, and target applications.

The other related survey papers focus on IM for FL~\cite{SP_MD_IEEE, SP_MD_ARXIV, SP_FL_TECHRXIV, SP_FL_MD}.
Zhan~\etal survey the IM design dedicated to FL~\cite{SP_MD_IEEE}. They summarize the state-of-the-art research efforts on the measures of clients' contribution, reputation, and resource allocation in FL. Zeng~\etal also survey the IM design for FL~\cite{SP_MD_ARXIV}. However, in this publication, the authors focus on IM such as Shapley values, the Stackelberg game, auction, context theory, and reinforcement learning. Besides~\cite{SP_MD_IEEE} and~\cite{SP_MD_ARXIV}, Ali~\etal~\cite{SP_FL_TECHRXIV} also summarize involved actors (e.g., number of publishers and workers), evaluation datasets as well as advantages and disadvantages of the mechanisms and security considerations. Tu~\etal~\cite{SP_FL_MD} provide a comprehensive review of economic and game theoretic approaches to incentivize data owners to participate in FL. In particular, they cluster applications of Stackelberg games, non-cooperative games,  sealed-bid auction models, reverse action models as well as contract and matching theory for incentive MD in FL.
Nguyen~\etal investigate opportunities and challenges of BC-based FL in edge computing~\cite{FLBlockchainOpportunitiesChallanges}. Finally, Wang~\etal surveyed BC-based FLF with a particular focus on FLF system compositions \cite{BlockchainFLSurvey}.


As revealed from our analysis and summarized in \tablename~\ref{table:comparison_of_related_survey_papers}, the existing survey papers lack a holistic analysis of FL frameworks that are both decentralized \textit{and} incentivized. Such a review is urgently needed since the simultaneous implementation of these features comes with additional interdisciplinary challenges while being crucial to establishing a fair and trustworthy FL framework to the benefit of the data owner. 
To fill this research gap, this paper provides the first systematic literature review on the topic of BC-enabled decentralized FL with IM.


%% file: 4_results.tex
\section{Systematic Literature Review}\label{sec:results}
The goal of the systematic literature review is the identification of decentralized collaborative learning solutions where participation is rewarded. For that, relevant publications are retrieved, filtered, and analyzed following a methodical, reproducible procedure. The procedure is inspired by the PRISMA methodology~\cite{moher2015preferred} and augmented with the guide for information systems proposed by Okoli~\etal and Kitchenham~\etal~\cite{okoli2010guide, kitchenham2004procedures}. The five core steps of the systematic approach include \one defining research questions, \two searching for literature, \three screening, \four reviewing, \five selecting and documenting relevant publications, and extracting relevant information. 
\par A detailed visualization of the search and screening step can be found in Figure~\ref{fig:tikz:prisma-diagram}. Details about the SLR are outlined in the Appendix. 

\begin{figure}[t]
    \centering
    \resizebox{\columnwidth}{!}{%
        \input{PRISMA_grafik_org}
    }
    \caption{Flow diagram of the search and screening step in the PRISMA methodology.}
    \label{fig:tikz:prisma-diagram}
\end{figure}
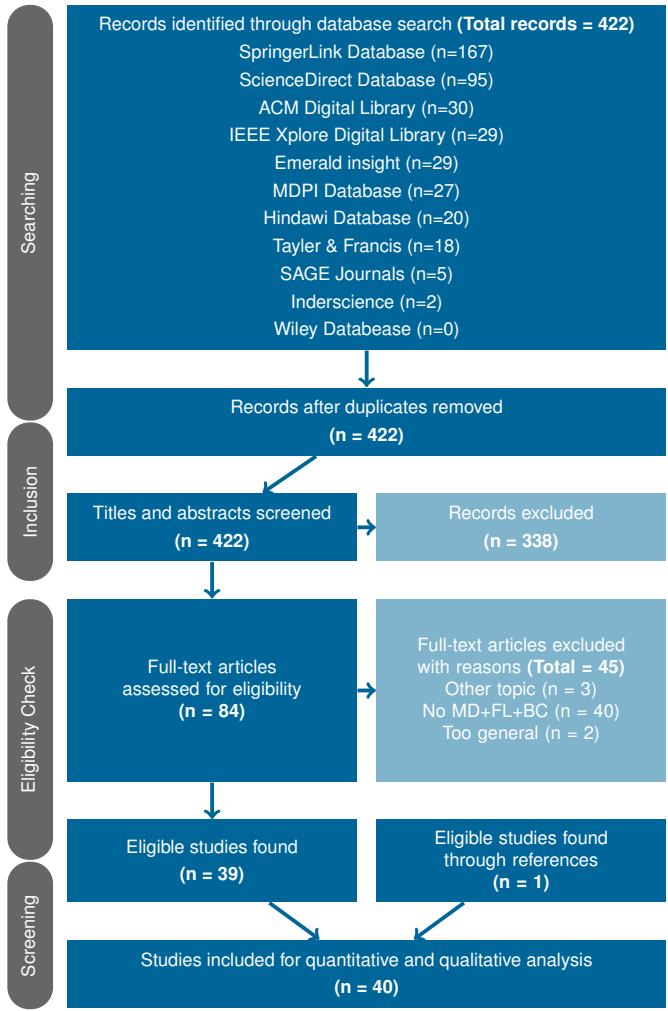

Subsections \ref{RQ1} to \ref{RQ5} systematically present the results of the literature review by answering our five research questions:

\begin{enumerate}
    \item[RQ1] Overview: \one What are the possible applications of FLF? \two What problems were solved? 
    
    \item[RQ2] BC: \one What is the underlying BC architecture? \two How is BC applied within the FLF and what operations are performed? \three Is scalability considered? 

    \item[RQ3] IM: \one How are IMs analyzed? \two How are the contributions of workers measured?
    
    \item[RQ4] FL: \one Is the performance of the framework reported? \two How comprehensive are the experiments? \three Are non-IID scenarios simulated? \four Are additional privacy methods applied? \five Is the framework robust against malicious participants?
    
    \item[RQ5] Summary: What are the lessons learned from the review?
\end{enumerate}

Each subsection is complemented by an explanatory table that classifies the considered papers according to categories defined in \tablename~\ref{tab:columns_definition}. Hereafter, we use n.a. (not applicable) for the items that are not applicable. Likewise, we use n.s. (not stated) for the items that should be stated but are not. We also use \checkmark\ for the items that satisfy a condition while leaving cells empty when they do not.
\begin{table*}[t]
    \centering
    \caption{Definition of columns in the overview tables.}
    \label{tab:columns_definition}
    \rowcolors{2}{tableoddrow}{tableevenrow}
    \resizebox{\linewidth}{!}{
    \begin{tabular}{C|LLL}
         \rowcolor{tableheader}
         Table & Column & Definition & Examples \\
         \hline
         \cellcolor{tableoddrow} & Application & Fields of applications & Generic, IoT \\
         \cellcolor{tableoddrow} & Setting & Whether a setting of FL is given & CS, CD\\
         \cellcolor{tableoddrow} & Actors & Actors assumed in the FLF & Workers \\
         \cellcolor{tableoddrow} & Setup & Whether how a system is set up was given (e.g., who deploys a BC) & \checkmark\\
         \cellcolor{tableoddrow} & \cellcolor{tableevenrow} & To which domain each work contributes &  \\
         \cellcolor{tableoddrow} & \cellcolor{tableevenrow} & \hspace{1em} SPoF: Single point of failure & \checkmark \\
         \cellcolor{tableoddrow} & \cellcolor{tableevenrow} & \hspace{1em} BC: Blockchain & \checkmark \\
         \cellcolor{tableoddrow} & \cellcolor{tableevenrow} & \hspace{1em} FL: Federated learning & \checkmark \\
         \cellcolor{tableoddrow} & \cellcolor{tableevenrow} & \hspace{1em} IM: Incentive mechanisms & \checkmark \\
         \cellcolor{tableoddrow} & \cellcolor{tableevenrow} & \hspace{1em} CM: Contribution measurement & \checkmark \\
         \multirow{-11}{*}{\cellcolor{tableoddrow}FL (\tablename~\ref{table:FL})} & \multirow{-7}{*}{\cellcolor{tableevenrow}Domains} & \hspace{1em} SP: Security \& privacy & \checkmark \\
         \hline
         \cellcolor{tableevenrow} & \cellcolor{tableoddrow} & Whether the following operations are executed on the BC & \\
         \cellcolor{tableevenrow} & \cellcolor{tableoddrow} & \hspace{1em} Agg.: Model aggregation & \checkmark \\
         \cellcolor{tableevenrow} & \cellcolor{tableoddrow} & \hspace{1em} Cor.: Coordination & \checkmark \\
         \cellcolor{tableevenrow} & \cellcolor{tableoddrow} & \hspace{1em} Pay.: Payment & \checkmark \\
         \cellcolor{tableevenrow} & \multirow{-5}{*}{\cellcolor{tableoddrow}Operations} & \hspace{1em} Str: Storage & \checkmark \\
         \cellcolor{tableevenrow} & BC & BC used in the FLF & Ethereum \\
         \cellcolor{tableevenrow} & Consensus & Consensus mechanism used & PoW \\
         \cellcolor{tableevenrow} & \cellcolor{tableevenrow} & Items: Types of data stored on-chain & \checkmark \\
         \cellcolor{tableevenrow} & \multirow{-2}{*}{\cellcolor{tableevenrow}On-chain} & Eval.: Whether the on-chain storage amount was evaluated & \checkmark \\
         \cellcolor{tableevenrow} & \cellcolor{tableoddrow}Off-chain & \cellcolor{tableoddrow}Whether off-chain storage (e.g., IPFS) is used & \cellcolor{tableoddrow}\checkmark \\
         \multirow{-11}{*}{\cellcolor{tableevenrow}BC (\tablename~\ref{table:blockchain})} & \cellcolor{tableevenrow}Scalability & \cellcolor{tableevenrow}Whether scalability is considered & \cellcolor{tableevenrow}\checkmark \\
         \hline
         \cellcolor{tableoddrow} & Sim. & Whether a game was evaluated via simulation & \checkmark \\
         \cellcolor{tableoddrow} & Theoretical analysis & Whether a game was theoretically analyzed & \checkmark(Contract theory) \\
         
         \cellcolor{tableoddrow} & \cellcolor{tableevenrow} & Costs: Costs considered in analysis & Energy \\
         \cellcolor{tableoddrow} & \cellcolor{tableevenrow} & Metrics: Metrics to validate workers' contribution & Accuracy \\
         \cellcolor{tableoddrow} & \cellcolor{tableevenrow} & Abs.: Whether the metric is absolute & Accuracy \\
         \cellcolor{tableoddrow} & \cellcolor{tableevenrow} & Rel.: Whether the metric is relative & Accuracy \\
         \cellcolor{tableoddrow} & \cellcolor{tableevenrow} & Rep.: Whether reputation is considered & Accuracy \\
         \multirow{-8}{*}{\cellcolor{tableoddrow}IM (\tablename~\ref{table:incentive_mechanism})} & \multirow{-6}{*}{\cellcolor{tableevenrow}Contribution measurement} & Validator: Actors that validate workers' contribution & Task requesters \\
         \hline
         \cellcolor{tableevenrow} & \cellcolor{tableevenrow} & Types of ML tasks & \\
         \cellcolor{tableevenrow} & \cellcolor{tableevenrow} & \hspace{1em}Clf: Classification & \checkmark \\
         \cellcolor{tableevenrow} & \cellcolor{tableevenrow}\multirow{-3}{*}{Tasks} & \hspace{1em}Rgr: Regression & \checkmark \\
         \cellcolor{tableevenrow} & Datasets & Datasets used for evaluation & MNIST \\
         \cellcolor{tableevenrow} & \#Clients & Number of clients in the experiments & 10 \\
         \cellcolor{tableevenrow} & Algorithms & FL algorithms used & FedAvg \\
         \cellcolor{tableevenrow} & Privacy & Whether privacy protection methods were applied & \checkmark \\
         \cellcolor{tableevenrow} & Non-IID & Whether the non-IID condition was assumed & \checkmark \\
         \cellcolor{tableevenrow}& \cellcolor{tableevenrow} & Whether security analysis was given against the following attacks & \\
         \cellcolor{tableevenrow}& \cellcolor{tableevenrow} & \hspace{1em}BT: Blockchain tampering &  \checkmark \\
         \cellcolor{tableevenrow}& \cellcolor{tableevenrow} & \hspace{1em}RP: Random model poisoning &  \checkmark \\
         \cellcolor{tableevenrow}& \cellcolor{tableevenrow} & \hspace{1em}RT: Reputation tampering &  \checkmark \\
         \cellcolor{tableevenrow}& \cellcolor{tableevenrow}\multirow{-5}{*}{Adversaries} & \hspace{1em}SP: Systematic model poisoning &  \checkmark \\
         \cellcolor{tableevenrow} & Imp. & Whether the BC part was implemented for evaluation & \checkmark \\
         \multirow{-15}{*}{\cellcolor{tableevenrow}Experiments (\tablename~\ref{table:experiments})} & Per. & Whether the performance of FL models was measured & \checkmark \\
         \hline
    \end{tabular}
    }
\end{table*}

\subsection{RQ 1: Overview}\label{RQ1}
\begin{table*}[t]
    \centering
    \caption{Overview of decentral and incentivized FL frameworks.}
    \label{table:FL}
    \resizebox{\textwidth}{!}{
        \input{table_overview}
    }
\end{table*}

%

\subsubsection{RQ 1-1: What are possible applications of FLF?}
\label{sec:RQ1_applications}
Table~\ref{table:FL} shows the summary of FL frameworks. Although most of the surveyed papers do not target specific applications (28 out of 40) due to the generalizability of neural networks, some are dedicated to specific applications, namely IoT (6 out of 40), Internet of Vehicles (IoV) (5 out of 40) and Finance (1 out of 40). 
Applications of FLF in special domains may require additional constraints and characteristics. The heterogeneity of the required properties across those domains leads to vast differences in the design choices of function, operations, and storage of BC, contribution measurement, and privacy requirements. 

One of the major application scenarios is IoT (e.g., \cite{LW4_FL_IIoT, LW5_FL_HomeAppliances_IoT, MH5_Lei2021}). Sensor-equipped devices collect environmental information and execute model updates thanks to advances in neural engines while edge servers are often assumed to aggregate models that are trained by local sensor devices. For instance, power consumption measured at smart homes can be used for training an AI model of energy demand forecast~\cite{MH14_Rathore2019}. Zhao~\etal propose a FLF for smart home appliance manufacturers to obtain AI models trained with their customers' usage information~\cite{LW5_FL_HomeAppliances_IoT}.

Some solved issues pertaining to the IoT-based FL~\cite{LW4_FL_IIoT, MH10_Qu2021}. Zhang~\etal propose a FL-based failure device detection method that takes into account the fact that sensor readings are often imbalanced since sensors are, in general, not deployed uniformly in a sensing area \cite{LW4_FL_IIoT}. They propose a modified FedAvg algorithm called centroid distance weighted federated averaging (\texttt{CDW\_FedAvg}) to obtain accurate models when local datasets are imbalanced at the devices. As sensor devices may not have enough resources to solely train neural networks, it is important to determine whether to delegate computationally intensive tasks to edge servers. Qu~\etal propose an algorithm to determine whether to offload computation to edge servers when communications between IoT devices and edge computers are unreliable \cite{MH10_Qu2021}. Beyond, Liu~\etal~\cite{Liu20206G} reflect on FL in the context of 6G communication and how both technologies are expected to empower each other. The review pinpoints communication cost, security, privacy, and training interference as the key challenges of FL and 6G communication.

FL is beneficial to many scenarios in ITS or IoV, e.g., optimized routing, congestion control, and object detection for autonomous driving (\cite{LW8_Privacy_IoV, MH1_Chai2021, MH6_Kansra2022, KT04_Zou2021WCNC, KT06_Wang2021TNSE}). Vehicles collect local information and train local models with collected data. Models are often aggregated by devices called Road Side Units (RSUs) and Mobile Edge Computers (MECs) which are often deployed on the road. In IoV, the CD setting is often preferred as mostly the same types of sensors are used to measure road conditions, and thus the common neural network model structure is shared by vehicles. As different locations have different road conditions, users need locally-optimized models, and thus scalability is a key issue. Furthermore, we need extra protection for users' location privacy. Zou~\etal propose a FLF for a knowledge trading marketplace where vehicles can buy and sell models that vary geographically~\cite{KT04_Zou2021WCNC}. Chai~\etal propose multiple BCs to deal with geographically dependent models~\cite{MH1_Chai2021}. Kansra~\etal integrate data augmentation, a technique to synthetically generate data such as images, into FL to increase model accuracy for ITS such as autonomous driving and road object identification~\cite{MH6_Kansra2022}. Wang~\etal propose a FLF dedicated to the crowdsensing of Unmanned Aerial Vehicles (UAVs)~\cite{KT06_Wang2021TNSE}. As UAVs are often equipped with multiple sensors and can be easily deployed to sensing areas, a FLF with UAVs has a huge benefit for ITS applications such as traffic monitoring and public surveillance.

Finance is the other domain that we found in the surveyed papers. He \etal proposed a FLF for commercial banks to better utilize customers' financial information~\cite{MH9_He2021}. Financial information such as credit level, risk appetite, solvency, movable and real estate owned are crucial sources to understanding the characteristics of customers of financial services. However, it is too sensitive to directly use them for data mining. Hence, a FLF is a viable framework for financial information management.

\subsubsection{RQ 1-2: What problems were solved?}
The problems solved by the papers can be categorized into \one the SPoF issue in FL, \two BC-related issues, \three lack of clients' motivation, \four how to fairly evaluate clients' contribution, \five security and privacy issues.

Most of the papers (29 out of 40) propose a system architecture of FLF to solve the problems of a SPoF in the current centralized server-clients architecture. More specifically, this issue is rooted in the structure of the original FL where an aggregation server is collecting local model updates from clients in a centralized manner. The idea to mitigate this issue is to decentralize the processes involved in FL using BC technologies. Each paper proposes operations, functions, and protocols processed in and outside the smart contract. Furthermore, some solve the issue of scalability in the FLF (e.g.,~\cite{MH1_Chai2021, MH5_Lei2021}) and BC-related issues such as energy waste of consensus algorithms (e.g.,~\cite{KT08_Qu2021TPDS, KT09_Zhang2021IJHPSA}). We will go into the proposed system architectures and BC-related issues in Section~\ref{sec:results_blockchain}. 

An incentive mechanism is integrated into FLFs to solve the problem of a lack of client motivation. The basic idea is to give monetary incentives to clients in return for their effort in training a local model. The incentive mechanism is also leveraged to solve the model poisoning attack which is an attack on a model update to deteriorate the quality of a global model by malicious clients' providing bogus local model updates. The idea for demotivating such attacks is to devise an incentive mechanism that penalizes malicious activities. Furthermore, a reputation score based on contribution is also useful to screen potentially malicious clients. Here, we need a contribution measurement metric to fairly evaluate the quality of clients' model updates and detect the attacks. Details about the proposed incentive mechanisms and contribution measurements will be covered in Section~\ref{sec:results_incentive_mechanism}.

20 out of 40 papers propose approaches to solve issues related to security and privacy. With few exceptions (i.e., attacks on reputation~\cite{MH2_Kang2019, MH7_Mugunthan2022} and~\cite{KT09_Zhang2021IJHPSA}), both security and privacy issues are rooted in local model updates. The security issue is related to the model poisoning which we mentioned above, while the privacy issue is related to sensitive information that might be leaked from the local updates. We will further summarize the works that solve the security and privacy issues in Section~\ref{sec:results_experiments}.

\subsection{RQ 2: Blockchain}
\label{sec:results_blockchain}
\begin{table*}[t]
    \centering
    \caption{Overview of BC features.}
    \label{table:blockchain}
    \resizebox{\textwidth}{!}{
        \input{table_blockchain}
    }
\end{table*}
\subsubsection{RQ 2-1: What is the underlying BC architecture?}
Table~\ref{table:blockchain} shows the overview of BC features.
The BC system and its underlying consensus mechanism are an influential part of the FLFs infrastructure. FLFs are heterogeneous in terms of architecture, operation and storage requirements, contribution calculation, actors, and applied cryptography. Customized and tailored BC solutions may be required with respect to the underlying use case. Due to its restrictive scalability in terms of computation and storage, most of the analyzed FLFs apply BC as a complementary element in a more complex system, with a few exceptions \cite{LW1_witt2021rewardbased,LW10_refiner, MH11_Desai2021, KT01_Toyoda2019BigData, KT02_Toyoda2020Access}. BC systems themselves are complex distributed systems, heterogeneous across many dimensions, yet can roughly be categorized into public, private, and permissioned BCs. 
\begin{enumerate}
    \item \textbf{Public}: BCs are open access where participants can deploy contracts pseudo-anonymously
    \item \textbf{Private}: BCs do not allow access for clients outside the private network and require an entity that controls who is permitted to participate
    \item \textbf{Permissioned}: BCs are private BCs with a decentralized committee that controls the onboarding process
\end{enumerate}
Note that the FLFs that utilize open-source public BCs such as Ethereum \cite{LW3_Kumar2020, LW4_FL_IIoT, LW6_Rahmadika2020, LW10_refiner, LW12_Rahmadika2021_unlinkable, LW14_5G_Rahmadika2021,MH7_Mugunthan2022, MH11_Desai2021, MH13_Li2020, MH14_Rathore2019, KT07_Ur_Rehman2020INFOCOMW}, Stellar \cite{MH8_Fadaeddini2019} and EOS \cite{KT03_Martinez2019CyberC} were not deployed on the respective public BC in the experiments due to the enormous costs this would incur. Hyperledger Fabric \cite{HyperledgerFabric} or Corda \cite{corda} are permissioned BCs running on private networks, allowing for faster throughput through a limited amount of potential nodes. This makes these frameworks more suitable for applications where BC replaces computationally expensive operations such as aggregation or storage of neural network models. 

The consensus protocol ensures the alignment and finality of a version across all distributed nodes without the need for a central coordination entity. While PoW is the most common mechanism applied in Bitcoin and Ethereum, it comes at the cost of wasting computational power on brute-forcing algorithmic hash calculations for the sole purpose of securing the network. Since many operations within the FLF frameworks are computationally expensive, these tasks can be integrated into the consensus mechanism which creates synergy and might be a better use of resources. Examples of consensus mechanisms can be found where the model accuracy is verified (Proof-of-Knowledge \cite{MH1_Chai2021}), reputation scores are checked (Proof-of-Reputation \cite{LW7_Zhang_Reputation}), the model parameters are securely verified (Proof-of-Federated-Learning \cite{KT08_Qu2021TPDS}), the Shapley value is calculated for contribution measurement \cite{LW9_FedCoin} or verification of capitalizing on efficient AI hardware (Proof-of-Model-Compression \cite{KT09_Zhang2021IJHPSA}).

\subsubsection{RQ 2-2: How is BC applied within the FLF and what operations are performed?}
BC technology is applied to mitigate the single point of failure and power imbalance of the server-worker topology of traditional FL through a transparent, immutable, and predictable distributed ledger. Embedded cryptocurrencies suit the useful property of real-time reward payments for predefined actions at the same time. In general, Turing-complete smart-contract-enabled BCs allow for a variety of possible complementary features for the FL training process, namely aggregation, payment, coordination, and storage: 
\begin{enumerate}
    \item \textbf{Aggregation}: The aggregation of model parameters, can be performed by a smart contract on top of BC \cite{LW11_RewardResponseGame, LW2_Weng2021, MH5_Lei2021, MH9_He2021, KT01_Toyoda2019BigData, KT02_Toyoda2020Access}. Since BC is assumed to be failure resistant, this strengthens the robustness against possible single-point of failure of an aggregation server. In addition, the deterministic and transparent rules of smart contracts ensure inherent trust with an equal power distribution among participants, while the transparency ensures auditability of contributions. Yet since every node in the BC has to compute and store all information, submitting a model to the smart contract for aggregation causes overhead in terms of both computation and storage on the BC. Assuming $n$ FL-workers and $m$ BC nodes over $t$ rounds, the BC scales with $\mathcal{O}(t*n*m)$ which questions the feasibility of on-chain aggregation. 
    
    There are two papers that try to reduce data size for on-chain aggregation.
    Witt~\etal~\cite{LW1_witt2021rewardbased} proposed a system where 1-bit compressed soft-logits are stored and aggregated on the BC saving communication, storage, and computation costs by orders of magnitude. 
 
    Feng~\etal~\cite{MH5_Lei2021} employ a framework based on two BC layers where the aggregation process is outsourced to a mobile edge server.

    \item \textbf{Coordination}: Applying BC to coordinate and navigate the FL process allows for decentralization without the heavy on-chain overhead.
    
    Instead of aggregating the model on-chain, letting the BC choose a leader randomly can ensure decentralization \cite{LW5_FL_HomeAppliances_IoT,MH11_Desai2021, MH1_Chai2021, KT11_Xuan2021SCN}. Another way BC coordinates the FL process is by enabling the infrastructure for trustless voting atop the BC. Voting on the next leader (aggregator) \cite{MH8_Fadaeddini2019, MH9_He2021} or on each other's contributions \cite{KT01_Toyoda2019BigData,KT02_Toyoda2020Access, KT11_Xuan2021SCN} further democratizes the process. Beyond explicit coordination operations like voting or leader selection, the implicit function of storing crucial information and data for the FL process, \cite{LW10_refiner, MH13_Li2020, KT04_Zou2021WCNC, LW1_witt2021rewardbased}, verifying the correctness of updates \cite{MH11_Desai2021, LW5_FL_HomeAppliances_IoT} or keeping the registry of active members \cite{LW1_witt2021rewardbased, LW10_refiner, KT01_Toyoda2019BigData, KT02_Toyoda2020Access, LW2_Weng2021, LW9_FedCoin} is crucial for the FL workflow and implies coordination through BC as an always accessible, verifiable, transparent and immutable infrastructure.
    

    
    \item \textbf{Payment}:
    Many general-purpose BC systems include cryptocurrency capabilities and therefore allow for the incorporation of instant, automatic and deterministic payment schemes defined by the smart contract's business logic. This advantage was capitalized on by 26 of the 40 FLF we analyzed.
    
     
 Section~\ref{sec:results_incentive_mechanism} discusses the details of applied payment schemes in the context of reward mechanisms and game theory.
    \item \textbf{Storage}:
    Decentralized and publicly verifiable storage on the BC facilitates auditability and trust among participants. Even though expansive, since all BC nodes store the same information in a redundant fashion, it might make sense to capitalize on the immutability and transparency feature of BC and store information where either shared access among participants is required or where verifiability of the history is required to hold agents accountable for posterior reward calculations \cite{LW7_Zhang_Reputation}. In particular, machine learning models \cite{LW2_Weng2021, LW8_Privacy_IoV,LW11_RewardResponseGame, MH1_Chai2021, Bao2019FLChain, MH4_Ma2021, MH9_He2021, MH10_Qu2021, MH11_Desai2021, KT01_Toyoda2019BigData, KT02_Toyoda2020Access}, 
    reputation scores \cite{LW7_Zhang_Reputation, LW13_TowardsReputationINFOCOMM}, User-information \cite{LW1_witt2021rewardbased, LW10_refiner, KT01_Toyoda2019BigData, KT02_Toyoda2020Access, LW2_Weng2021, LW9_FedCoin} and Votes \cite{KT01_Toyoda2019BigData, KT02_Toyoda2020Access, LW1_witt2021rewardbased} are stored on-chain of the respective FLF.

\end{enumerate}  

\subsubsection{RQ 2-3: Is the framework scalable?}
Especially if the FLF is intended to be used with hundreds to millions of devices, the scalability of the framework is an important characteristic. In particular, \one storing large amounts of data such as model parameters and \two running expansive computations on the BC e.g., aggregating millions of parameters, calculating expansive contribution measurements like Shapley Value or privacy-preserving methods hinder the framework to scaling beyond a small group of entrusted entities towards mass adoption. To overcome the scalability bottleneck of storage, some FLF applied an Interplanetary File System (IPFS) \cite{ipfs}, where data is stored off-chain in a distributed file system, using the content address as a unique pointer to each file in a global namespace over computing devices \cite{LW3_Kumar2020, LW5_FL_HomeAppliances_IoT, LW10_refiner, LW13_TowardsReputationINFOCOMM, MH2_Kang2019, MH7_Mugunthan2022, MH8_Fadaeddini2019, KT03_Martinez2019CyberC, KT07_Ur_Rehman2020INFOCOMW}. Other FLFs are based on novel design choices to tackle the scalability issues: Witt~\etal~\cite{LW1_witt2021rewardbased} applied compressed Federated Knowledge Distillation, storing only 1-bit compressed soft-logits on-chain. Chai~\etal~\cite{MH1_Chai2021} design a hierarchical FLF with two BC layers to reduce the computational overhead by outsourcing computation and storage to an application-specific sub-chain. Similarly, Feng~\etal~\cite{MH5_Lei2021} propose a two-layered design, where the transaction efficiency of the global chain is improved through sharding. Bao~\etal~\cite{Bao2019FLChain} employ an adaption of Counting Bloom Filters (CBF) to speed up BC queries in the verification step of their FLF. Desai~\etal~\cite{MH11_Desai2021} combine public and private BCs, with the former storing reputation scores for accountability and the latter used for heavy computation and storage. Furthermore, the authors apply parameter compression for further scalability improvements.

\subsection{RQ 3: Incentive Mechanisms}
\label{sec:results_incentive_mechanism}
\begin{table*}[t]
    \centering
    \caption{Overview of incentive mechanisms and contribution measurement.}
    \label{table:incentive_mechanism}
    \resizebox{\textwidth}{!}{
        \input{table_incentive_mechanism}
    }
\end{table*}

\subsubsection{RQ 3-1: How are incentive mechanisms analyzed?}
In general, the analysis comprises three steps: The first step is to determine what entities' behavior is examined. In FL, such entities could be workers or task requesters. The second step is to model the entities' utilities or profits. They can be obtained by taking the expectation of possible profits and costs into account. The last step is to analyze the defined utilities or profits. This could be done in a theoretical manner and/or via simulation. 30\% (12 out of 40) of the surveyed papers analyzed the incentive mechanism theoretically, while 45\% (18 out of 40 papers) measured workers' rewards via simulation. However, we only focus on the papers with theoretical analysis in this section as we do not see much technical depth or differences in the simulation-based analysis.

In the first step, 28 papers assume that only workers exist, while 12 FLFs have additional entities that pay rewards (e.g., task requesters) \cite{LW7_Zhang_Reputation, MH2_Kang2019, Bao2019FLChain, MH12_Zhu2021, KT01_Toyoda2019BigData, KT02_Toyoda2020Access, KT03_Martinez2019CyberC, KT05_Hu2021IoTJ, KT07_Ur_Rehman2020INFOCOMW, KT08_Qu2021TPDS, KT10_Gao2021ICPP, KT11_Xuan2021SCN}. In such a case, the utilities of both entities have to be analyzed such that task requesters can be profitable even if they pay rewards to workers. For instance, workers and task requesters are assumed in \cite{KT01_Toyoda2019BigData}, and it is vital to determine their behavioral assumptions (e.g., their goals, rationality, etc.).

The second step is to define the utilities or profits of entities. A utility is a one-dimensional measurable unit that quantifies an entity's value on an outcome and can have positive (e.g., rewards for workers and a value of AI models for task requesters) and negative values (e.g., a computation cost for workers and a total amount of payout for task requesters). Utilities and profits can be derived by subtracting costs from payouts (Eq.~\eqref{eq:utilityfunction}). Although the elements of payouts $\Pi$ are mostly straightforward (e.g., rewards for work contribution), the cost elements $C$ are dependent on the assumed application scenarios. Typical costs in the surveyed papers are computation, electricity (e.g., \cite{LW11_RewardResponseGame, MH1_Chai2021, MH2_Kang2019, KT01_Toyoda2019BigData, KT02_Toyoda2020Access, KT06_Wang2021TNSE}), data acquisition (e.g., pictures and sensor readings \cite{KT04_Zou2021WCNC, KT05_Hu2021IoTJ, KT06_Wang2021TNSE}) and privacy leakage due to model updates (e.g., \cite{KT05_Hu2021IoTJ, KT06_Wang2021TNSE, KT08_Qu2021TPDS}). Even multiple cost factors can be considered (e.g., \cite{KT04_Zou2021WCNC, KT05_Hu2021IoTJ, KT06_Wang2021TNSE}). 

The last step is to analyze the defined utilities and profits to ensure the robustness of the incentive mechanisms and derive the optimal reward allocation. A simple yet crucial analysis would be to prove that it is worthwhile for workers to join a FL task by showing that their profits are non-negative. For instance, Bao~\etal modeled requesters' and/or workers' profits with given rewards and costs and proved that their profits are non-negative \cite{Bao2019FLChain}. Utilities can be used to derive task requesters' and workers' optimal strategies by finding a point where utilities are maximized. By proving the existence of such a point, an equilibrium can be derived, which is a condition where entities (e.g., workers) cannot be better off deviating from their optimal strategies. An equilibrium state, if existent, is proof that a designed mechanism is stable. An equilibrium can be found by deriving the first- and second-order derivatives of the utility function with respect to the variable in question. For instance, Toyoda~\etal optimizes the workers' data size  \cite{KT02_Toyoda2020Access}. 

Other works that determine optimal prices for tasks: Wang~\etal propose a Q-learning-based approach to determine the optimal prices so that utilities are maximized via iterative learning processes \cite{KT06_Wang2021TNSE}. Similarly, Zou~\etal derive the optimal prices for workers with first- and second-order conditions when the value of data, transmission quality, and communication delay are the factors to determine their competitiveness and costs \cite{KT04_Zou2021WCNC}. Hu~\etal propose a two-stage optimization method to determine the optimal values of data and their prices in order by solving an Euler-Lagrange equation of their utilities. These kinds of two-stage optimization games are often formalized as a Stackelberg game \cite{LW11_RewardResponseGame, MH1_Chai2021}. Jian and Wu propose a Stackelberg-game-based incentive mechanism for FL \cite{LW11_RewardResponseGame}. They analyzed the equilibria of two reward policies where a contribution is measured by data size or accuracy by modeling an aggregation server as leader and workers as followers. The uniqueness of their analysis is that they incorporate training and uploading time into the analysis in terms of a constraint as each round has a deadline in the FL.

Chai~\etal propose a multi-leader multi-follower Stackelberg game to analyze their incentive mechanism in IoV \cite{MH1_Chai2021}. Aggregation servers (RSUs) are leaders while workers (vehicles) are followers, and aggregation servers first suggest prices and workers determine how much data they should collect and use for training so that both entities' utilities are maximized in order. Due to the high dimensionality of each worker's strategy, it is difficult to employ the traditional backward induction method to derive an equilibrium. Hence, they leverage the Alternating Direction Method of Multipliers (ADMM) algorithm~\cite{boyd2011distributed} to iteratively reach the social optimum point.

Three papers model the incentive mechanism in FL as a contract or contest: a task requester proposes a contract with a task description and its reward and workers can determine whether or not to sign such a contract and how many resources they will provide \cite{MH2_Kang2019}. A FL process can be also seen as a contest as workers need to work first, which incurs irreversible costs due to computation, whereas their rewards are not guaranteed at the time of model update submission. Toyoda~\etal give an incentive analysis based on the contest theory~\cite{KT01_Toyoda2019BigData, KT02_Toyoda2020Access}. Workers' utilities are used to derive how much effort workers' should exert on a task under the risk of not gaining prizes, while requesters' utility is used to determine how a prize should be split among workers.


\subsubsection{RQ 3-2: How are the contributions of clients measured?}
Incentive mechanisms require \one the measurement of contribution by each client to \two fairly distribute rewards. However, the clients' contributions in form of model updates or gradients do not imply direct information on the overall performance metric like the accuracy of the global model.


The metrics used in the literature can be categorized into absolute and relative ones. The absolute metrics are metrics that can be measured without others' local model updates. For instance, a loss function can be measured from a local model and a global model, and the difference between them can be used as a metric for contribution measurement. Although the majority of absolute metrics are based on the accuracy (e.g., \cite{LW3_Kumar2020, LW4_FL_IIoT}) and data size (e.g., \cite{LW6_Rahmadika2020, LW11_RewardResponseGame}), other factors are also proposed such as energy consumption (e.g., \cite{LW7_Zhang_Reputation, MH10_Qu2021}) and computation time \cite{KT12_Liu2021Sensors}. Some combine multiple metrics (e.g., \cite{LW7_Zhang_Reputation, MH13_Li2020}). Absolute metrics are generally straightforward but hard to validate, e.g., metrics that are based on the data size used depending on the client's honesty. In contrast, relative metrics can be measured by comparing submitted results (e.g., gradients, model updates) in terms of correlation or ranking. For instance, Zhao~\etal propose a metric based on the Euclidean distance of workers' model updates \cite{LW5_FL_HomeAppliances_IoT}. Likewise, Witt~\etal utilizes peer-truth serum \cite{Radanovic2016PTS} in the FL context, where contributions are measured based on the correlation of prediction on the labels of a public dataset \cite{LW1_witt2021rewardbased}. 
Voting is another approach to  determine clients' contribution relatively. For instance, clients choose the best model updates from the previous \texttt{FedAvg} round by ranking the respective updates based on the accuracy using their local datasets \cite{KT01_Toyoda2019BigData, KT02_Toyoda2020Access, KT11_Xuan2021SCN}.

A similar metric is clients' reputations. If the same clients are assumed to join different FL tasks, reputation scores calculated based on clients' past contributions can be used to determine the reward distribution (e.g., \cite{LW5_FL_HomeAppliances_IoT, LW7_Zhang_Reputation, MH2_Kang2019, MH13_Li2020}). For instance, Kang~\etal propose to calculate workers' reputation based on a direct opinion by a task requester and indirect opinions by other task requesters \cite{MH2_Kang2019}.
\newline
However, even if the clients' individual contribution is measured, the question regarding fair distribution remains an unsolved issue. The Shapley value is an approach to determine payouts to workers based on the marginal utility added, taking all possible combinations of contributors into consideration \cite{Shapley1953}. Three papers propose to use the Shapley value for fair reward distribution in the FL context \cite{LW9_FedCoin, MH9_He2021, MH4_Ma2021}. Liu~\etal applies the Shapley value based on the accuracy of a test dataset~\cite{LW9_FedCoin}. He~\etal compared their Shapley-value-based method with three approaches, namely \one equal distribution, \two a method based on individual contribution, and \three a method called the labor union game where only the order of submission is taken into account to contribution measurement, and found that the Shapley-value-based method outperforms the others in terms of workers' motivation and fairness \cite{MH9_He2021}. Ma~\etal propose a method to calculate Shapley values even if model updates are masked to preserve workers' privacy \cite{MH4_Ma2021}.

Which entities validate the contribution is an open issue, complementary to the contribution measurement. The validators in FLF can be classified into \one aggregation servers FL (e.g.~\cite{LW4_FL_IIoT, LW8_Privacy_IoV, MH1_Chai2021}), \two task requesters (e.g.~\cite{LW7_Zhang_Reputation, MH2_Kang2019, MH13_Li2020}), \three validators whose task is only to measure contribution (\cite{LW10_refiner, MH8_Fadaeddini2019}), \four BC nodes (e.g., \cite{LW3_Kumar2020, KT08_Qu2021TPDS, KT12_Liu2021Sensors}), \five workers (e.g., \cite{MH7_Mugunthan2022, KT10_Gao2021ICPP, KT11_Xuan2021SCN}) and \six smart contracts (e.g., \cite{LW1_witt2021rewardbased, LW6_Rahmadika2020, MH4_Ma2021}).
Some of the works assume that aggregation servers, task requesters, or validators are expected to possess datasets to calculate the metrics discussed above. As reviewed in Section \ref{sec:results_blockchain}, others propose custom BC architectures for FL where the validation process is integrated into the consensus mechanism, making BC nodes validators.
In some scenarios, aggregation servers, task requesters, and BC nodes take up the role of validators since they aggregate the model updates. However, datasets for validation may not be always available. Furthermore, metrics based on the correlation of predicted labels do not require any validation dataset and can even be measured in a smart contract~\cite{LW1_witt2021rewardbased}.

\subsection{RQ 4: Experiments}
\label{sec:results_experiments}
\begin{table*}[t]
    \centering
    \caption{Overview of experiments. \textit{\footnotesize (BCWD = Breast Cancer Wisconsin Data Set, BT = Blockchain Tampering, DGHV = Dijk-Gentry-Halevi-Vaikutanathan Algorithm, DP = Differential Privacy, ECC = Elliptic Curve Cryptography, HE = Homomorphic Encryption, HDD = Heart Disease Data Set, KDD = Knowledge Discovery and Data Mining Tools Competition, RP = Random Model Poisoning, RSA = Rivest-Shamir-Adleman Cryptosystem, RT = Reputation Tampering, SA = Secure Aggregation~\cite{MH_SECURE_AGG}, SP = Systematic Model Poisoning, ZKP = Zero-Knowledge Proof, 2PC = 2-Party Computation)}}
    \label{table:experiments}
    \resizebox{\textwidth}{!}{
        \input{table_experiments}
    }
\end{table*}
Conducting experiments is a key element of FLF development for two reasons. Firstly, the implementation of an example testifies to the feasibility of the approach and gives the authors the chance to identify weaknesses of their frameworks, e.g., poor scalability. Secondly, conducting experiments allows the comparison of the proposed approaches with each other, e.g., based on the accuracy of the models on standardized test sets. We screened the papers for experiments, and when present, examined them according to nine criteria (\tablename~\ref{table:experiments}).

\subsubsection{RQ 4-1: Is the performance of the framework reported?}
The large majority of papers report results of their experiments expressed in either loss, accuracy, or F1 score (84.8\%, 28 out of 33 papers with experiments). The remaining instead focus on the performance of their novel group-based Shapley value calculation for contribution measurement~\cite{MH4_Ma2021}, the user interface~\cite{LW9_FedCoin}, the computational effort and adversarial influence~\cite{MH11_Desai2021}, or game-theoretic quantities such as utility values and rewards~\cite{KT06_Wang2021TNSE}. We note that comparability of the approaches is not given through the conducted experiments, since even when using the same data sets and the same evaluation metrics, different experimental scenarios are investigated.
In conclusion, to obtain insightful results, experiments should compare the performance (accuracy and computational effort) of an incentivized, decentralized FL system in a standardized challenging environment (non-IID, adversaries) with either the performance of a traditional centralized FL system or the performance of a locally trained model without FL. Ideally, the effectiveness of IM and decentralization efforts are reflected in the FLF performance through a holistic experimental design.  

\subsubsection{RQ 4-2: How comprehensive are the experiments?}
First, it was found that the majority of publications do include experiments. Only seven of 40 papers did not conduct experiments~\cite{LW13_TowardsReputationINFOCOMM,MH6_Kansra2022,MH8_Fadaeddini2019,MH9_He2021,KT01_Toyoda2019BigData, KT02_Toyoda2020Access,KT07_Ur_Rehman2020INFOCOMW}. However, the analysis also shows that only 45\% of the experiments implement the actual BC processes (15 out of 33 papers with experiments). Instead, the distributed functionality was simulated or its impact estimated. For instance, Mugunthan~\etal~\cite{MH7_Mugunthan2022} focus on the evaluation of the frameworks' contribution scoring procedure by simulating collusion attacks on the FL procedure. The effect of introducing BC to the FL framework was accounted for by estimating the per-agent gas consumption. Similarly, Chai~\etal~\cite{MH1_Chai2021} conducted experiments specifically designed to investigate the Stackelberg game-based incentive mechanism. The authors accomplish this without implementing the BC processes.

To test the FL functionality of the framework, an ML problem and a dataset must be selected. For the ML application and the dataset used, we observe a high homogeneity. Almost all experiments realize classification problems and use publicly available benchmark datasets. The most common are MNIST (handwritten digits)~\cite{MH_MNIST} and its variations, as well as CIFAR-10 (objects and animals)~\cite{MH_CIFAR}. Only Rathore~\etal~\cite{MH14_Rathore2019} and Li~\etal~\cite{LW8_Privacy_IoV} did not perform classification tasks. Rathore~\etal~\cite{MH14_Rathore2019} performed object detection on the PASCAL VOC 12 dataset~\cite{MH_PASCAL}. Object detection typically combines regression and classification by predicting bounding boxes and labeling them. Li~\etal~\cite{LW8_Privacy_IoV} applied their FLF to autonomous driving and minimized the deviations in steering-wheel rotation between a human-driven and simulation-driven vehicle. This corresponds to a regression task.

As to the number of training data holders, the experiments considered between one~\cite{LW8_Privacy_IoV} and 900~\cite{MH12_Zhu2021} clients. In general, one would expect papers specifying cross-silo settings to test with fewer ($<$100~\cite{SP_MD_ARXIV}) and papers specifying cross-device settings to test with more ($>$100) clients. Of the frameworks clearly designed for a cross-device application, it is noticeable that only Kang~\etal~\cite{MH2_Kang2019} and Desai~\etal\cite{MH11_Desai2021} conduct experiments with 100 participants or more. On the contrary, Rahmadika~\etal~\cite{LW12_Rahmadika2021_unlinkable} test with as many as 100 participants, although only designing a cross-silo framework.

Regarding the FL algorithm, the classic FedAvg~\cite{BrendanMcMahan2017} is mainly used. Furthermore, in some experiments algorithms are used that mitigate the problem of catastrophic forgetting (Elastic Weight Consolidation (EWC)~\cite{LW3_Kumar2020}), reduce the communication overhead (Federated Knowledge Distillation (FD)~\cite{LW1_witt2021rewardbased}, signSGD~\cite{MH11_Desai2021}), or show more robust convergence for non-IID and other heterogeneous scenarios (\texttt{FedProx}~\cite{LW10_refiner}, Centroid Distance Weighted Federated Averaging (\texttt{CDW\_FedAvg})~\cite{LW4_FL_IIoT}). Chai~\etal~\cite{MH1_Chai2021} and Mugunthan~\etal~\cite{MH7_Mugunthan2022} design custom FL algorithms. Specifically, Chai~\etal~\cite{MH1_Chai2021} propose a FLF with two aggregation layers in order to promote scalability. In their FL algorithm, nodes in the middle layer aggregate the local model updates of associated nodes in the lowest layer. This semi-global model is then fine-tuned by the middle layer nodes based on data collected by the middle layer nodes themselves. Finally, nodes in the top layer aggregate the fine-tuned models from the middle layer nodes into a global model which is eventually passed back to the lowest layer nodes. All aggregations are weighted by the training dataset size. Mugunthan~\etal~\cite{MH7_Mugunthan2022} propose a FLF where all clients evaluate and score the differentially encrypted locally trained models of all other clients. These scores are reported to a smart contract which computes an overall score for each local model. Eventually, each client aggregates the global model from all local models, weighted by the overall score.

\subsubsection{RQ 4-3: Are non-IID scenarios simulated?}
\par In real-world applications of FL, the training data is often not Independent and Identically Distributed (non-IID) between the clients. This affects the performance of the global model and adds an additional layer of complexity with respect to contribution measurement. Hence, how we simulate non-IID scenarios with open datasets is crucial. In 11 publications and for various benchmark datasets, non-IID scenarios were considered. For the example of the MNIST dataset, Witt~\etal~\cite{LW1_witt2021rewardbased} simulate different levels of non-IID scenarios following the Dirichlet distribution as it can easily model the skewness of data distribution by varying a single parameter. Martinez~\etal~\cite{KT03_Martinez2019CyberC} split the dataset in overlapping fractions of various sizes, whereas Kumar~\etal~\cite{LW3_Kumar2020} divide the dataset so that each trainer only possesses data from two of the ten classes. In a less skewed setting, Kumar~\etal allocate data from at most four classes to each trainer, with each class being possessed by two devices. 

\subsubsection{RQ 4-4: Are additional privacy methods applied?}
\label{sec:results_experiments_privacy}
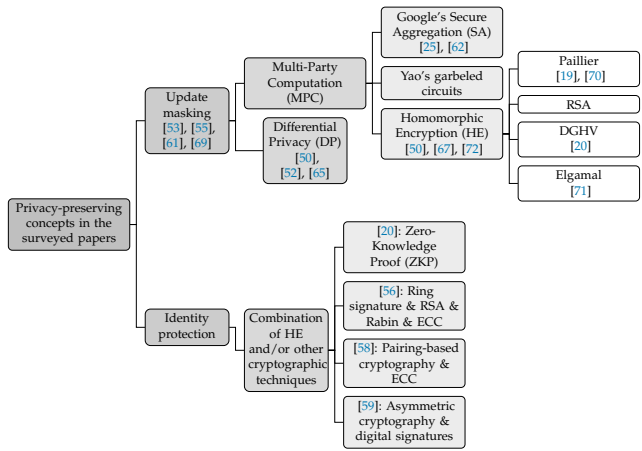
\begin{figure}
    \centering
    \resizebox*{0.95\linewidth}{!}{%
        \begin{forest}
            forked edges,
            for tree={
                grow'=east,
                draw,
                rounded corners,
                text width=2cm,
                node options={align=center},
            }       
            [Privacy-preserving \\concepts in the \\surveyed papers, fill=col5, parent, s sep=1cm
            [Update \\masking \\\cite{LW10_refiner, MH11_Desai2021, Bao2019FLChain, LW6_Rahmadika2020}, fill=col4
            [Multi-Party \\Computation (MPC), for tree={child, fill=col3}
            [Google's Secure \\Aggregation (SA) \\\cite{MH4_Ma2021, LW9_FedCoin}, for tree={child, fill=col2}]
            [Yao's garbeled \\ circuits, fill=col2]
            [Homomorphic \\Encryption (HE)\\\cite{MH9_He2021, MH14_Rathore2019, LW3_Kumar2020}, for tree={child, fill=col2}
            [Paillier \\\cite{MH12_Zhu2021, LW2_Weng2021}, fill=col1]
            [RSA, fill=col1]
            [DGHV\\\cite{LW8_Privacy_IoV}, fill=col1]
            [Elgamal \\\cite{MH13_Li2020},fill=col1]
            ]
            ]
            [Differential \\Privacy (DP) \\\cite{MH7_Mugunthan2022, LW5_FL_HomeAppliances_IoT, LW3_Kumar2020}, fill=col3]
            ]
            [Identity \\protection, fill=col4
            [Combination of HE and/or other cryptographic techniques, fill=col3
            [\cite{LW8_Privacy_IoV}: Zero-Knowledge \\Proof (ZKP) , for tree={child, fill=col2}]
            [\cite{LW12_Rahmadika2021_unlinkable}: Ring signature~\& RSA~\& Rabin~\& ECC , for tree={child, fill=col2}]
            [\cite{LW14_5G_Rahmadika2021}: Pairing-based cryptography~\& ECC , for tree={child, fill=col2}]
            [\cite{MH1_Chai2021}: Asymmetric cryptography~\& digital signatures , for tree={child, fill=col2}]
            ]
            ]
            ]
        \end{forest}
    }
    \caption{Privacy-preserving concepts employed in survey papers. \textit{\footnotesize (DGHV = Dijk-Gentry-Halevi-Vaikutanathan Algorithm, ECC = Elliptic Curve Cryptography, RSA = Rivest-Shamir-Adleman Cryptosystem. Papers with unspecified methods are added to next highest node.)}}
    \label{abb:PrivacyClassification}
\end{figure}

Even though FL's core objective is to maintain confidentiality through a privacy-by-design approach where model parameters are aggregated instead of training data, there remain innumerable attack surfaces \cite{FL_Attacks_Taxonomy}. Therefore, the presented frameworks employ additional privacy-preserving mechanisms which can be divided into two groups. \one Mechanisms that encrypt or obfuscate gradients and prevent malicious parties to draw conclusions about the data set. \two Mechanisms that hide the identity of participating parties. A classification of the employed privacy-preserving methods can be seen in \figurename~\ref{abb:PrivacyClassification}.

The methods of the first group can be further divided into \one approaches that are based on cryptographic secure MPC, and \two approaches that are based on Differential Privacy (DP). 

MPC refers to cryptographic methods by which multiple participants can jointly compute a function without having to reveal their respective input values to the other participants. MPC approaches include three groups of methods~\cite{MH_SECURE_AGG}. \one Google's Secure Aggregation (SA)~\cite{MH_SECURE_AGG} is specifically designed to achieve low communication and computation overhead and to be robust towards device dropout. It has been employed by Liu~\etal~\cite{LW9_FedCoin} and Ma~\etal~\cite{MH4_Ma2021}. Beyond implementing SA, the latter develops a group-based Shapley-value method for contribution measurement, since the native Shapley-value method cannot be applied to masked gradients. \two Yao's garbled circuits have not been applied to any of the analyzed frameworks, but are mentioned here for completeness. \three While Yao's garbled circuits were developed for 2-party secure computing, Homomorphic Encryption (HE) allows for higher numbers of participants~\cite{MH_SECURE_AGG}. As shown in \figurename~\ref{abb:PrivacyClassification}, HE has been employed in several works \cite{LW2_Weng2021}, \cite{LW8_Privacy_IoV}, \cite{MH9_He2021}, \cite{MH13_Li2020}, \cite{MH12_Zhu2021}, \cite{MH14_Rathore2019}. For that, different implementations of the homomorphic idea have been chosen, such as the Pallier cryptosystem, the Elgamal cryptosystem, or the Dijk-Gentry-Halevu-Vaikutanathan Algorithm (DGHV). Li~\etal~\cite{MH13_Li2020} choose the Elgamal cryptosystem that is less computationally expensive than other HE approaches. 

DP refers to a method where noise is drawn from a probability density function $p_{noise}(x)$ with expected value $\mathbb{E}(p_{noise}(x))=0$ obfuscates the individual contribution with minimal distortion of the aggregation. DP has been employed by Mugunthan~\etal~\cite{MH7_Mugunthan2022}, Zhao~\etal~\cite{LW5_FL_HomeAppliances_IoT}, and Kumar~\etal~\cite{LW3_Kumar2020}, with the latter combining the use of HE and DP. However, the fewer clients participate in the DP process, the heavier the distortion of the aggregated model, introducing a trade-off between privacy and model accuracy. Zhao~\etal~\cite{LW5_FL_HomeAppliances_IoT} mitigate the loss in accuracy by incorporating a novel normalization technique into their neural networks instead of using traditional batch normalization (e.g., \cite{MH7_Mugunthan2022, LW3_Kumar2020}). 
Besides MPC and DP, another technique for data set protection is chosen by Qu~\etal~\cite{KT08_Qu2021TPDS}. Instead of the clients sharing masked gradients, the FLF relies on requesters sharing masked datasets in the model verification step. This prevents other workers from copying the models while testing and evaluating them. HE and 2-Party Computation (2PC) are used. Zhang~\etal~\cite{LW10_refiner}, Desai~\etal~\cite{MH11_Desai2021}, Bao~\etal~\cite{Bao2019FLChain}, and Rahmadika~\etal~\cite{LW6_Rahmadika2020} also rely on the masking of gradients but do not specify the privacy-preserving mechanisms.

The second group of frameworks targets the protection of participants' identities through cryptographic mechanisms. For that, Rahmadika~\etal~\cite{LW12_Rahmadika2021_unlinkable} combine ring signatures, HE (RSA), Rabin algorithm, and Elliptic Curve Cryptography (ECC), while Chai~\etal~\cite{MH1_Chai2021} incorporate digital signatures and asymmetric cryptography approaches, and Rahmadika~\etal~\cite{LW14_5G_Rahmadika2021} perform authentication tasks through pairing-based cryptography and ECC. Only one framework implements measures for both masking gradients as well as hiding identities. Li~\etal~\cite{LW8_Privacy_IoV} use DGHV for masking gradients and Zero-Knowledge Proof (ZKP) for identity protection. 

Finally, He~\etal~\cite{MH9_He2021} specifically address the problem of aligning entities. This problem occurs in vertical federated learning where different parties hold complementary information about the same user. The parties have to find a way of matching this information without disclosing the identity of their users. To solve this problem, He~\etal employ Encrypted Entity Alignment which is a protocol for privacy-preserving inter-database operations~\cite{MH9_He2021}.     

\subsubsection{RQ 4-5: Is the framework robust against malicious participants?}
The experiments consider and simulate different types of adversaries, whereas some publications consider multiple types of attacks. Four groups of attack patterns were identified in the publications: random model poisoning, systematic model poisoning, reputation tampering (RT), and BC tampering. 
The most common attack considered in the experiments is random model poisoning. This includes attacks, where local models are trained on a randomly manipulated data set (\cite{LW10_refiner}, \cite{MH2_Kang2019}, \cite{MH7_Mugunthan2022}, \cite{KT10_Gao2021ICPP}) or where random parameter updates are reported (\cite{MH10_Qu2021}, \cite{LW1_witt2021rewardbased}, \cite{MH2_Kang2019}, \cite{LW5_FL_HomeAppliances_IoT}, \cite{MH1_Chai2021}, \cite{MH12_Zhu2021}, \cite{KT09_Zhang2021IJHPSA}). For instance, malicious agents in \cite{LW10_refiner} use a training dataset with intentionally shuffled labels, whereas in \cite{MH12_Zhu2021} the parameter updates are randomly perturbed with Gaussian noise. Kang~\etal~\cite{MH2_Kang2019} analyze the effects of a bad or manipulated data set by providing 8\% of the workers with training data where only a few classes are present, and another 2\% of the workers with mislabeled data. Kang~\etal quantify the insufficiency of the dataset using the earth mover's distance. 

The second most commonly simulated type of attack is systematic model poisoning where the attackers manipulate the model through well-planned misbehavior. In \cite{MH11_Desai2021}, a fraction of workers collude and manipulates their image classification data sets by introducing a so-called trojan pattern: the malicious agent introduces a white cross to a certain fraction of a class, e.g., to 50\% of all dog pictures in an animal classification task and re-labels these data points as horse pictures. This creates a backdoor in the model that cannot be detected by subjecting the model to dog or horse pictures which will be correctly classified. However, pictures with the trojan pattern will be misclassified. Other forms of systematic model poisoning can be found with Witt~\etal~\cite{LW1_witt2021rewardbased}, Mugunthan~\etal~\cite{MH7_Mugunthan2022}, Gao~\etal~\cite{KT10_Gao2021ICPP}.

The third type of attack that was simulated is Reputation Tampering (RT). Here, malicious agents intentionally provide colluding agents with perfect reputation or voting scores \cite{MH2_Kang2019,MH7_Mugunthan2022}. 
The fourth type of attack is BC tampering \cite{KT09_Zhang2021IJHPSA}. Here, malicious miners intentionally fork the BC and prevail by building a longer branch faster than the honest miners.

\subsection{RQ 5: Summary: What are Lessons Learned?}\label{RQ5}

The inherent complexity of FLF leads to heterogeneity of the  scientific research across the dimensions \one application, \two overall design, \three special focus on open issues, and \four details and thoroughness. 

\textbf{Application}: Although the majority of analyzed works offer application independent frameworks (classified as ``generic'' in Table~\ref{table:FL}) other FLFs are applied across IoT, Industrial-IoT (IIoT), IoV, and Finance. The heterogeneity of the required properties across those domains causes differences in the design choices of function, operations, storage of BC, contribution measurement, and privacy requirements.

\textbf{Variety of possible design choices}: In addition to the domain-specific influence on the system architecture, design choices about the FL algorithms, communication protocol, applications of BC within the ecosystem, BC technology (existing or novel), storage and operation on BC, security trade-offs, mechanism design, contribution measurement, etc. add to the complexity and overall variety of such systems. For example, some works apply BC as the outer complementary layer \cite{LW7_Zhang_Reputation} while BC is the core infrastructure for coordination, storage, aggregation, and payment in other FLFs \cite{LW1_witt2021rewardbased, LW10_refiner}. Furthermore, some works developed application-specific BC systems, while others tried to embed a FLF on top of existing BC frameworks such as Ethereum for cheaper and pragmatic deployment. Our survey exposes a similar variety in the choice of the contribution measurement: The spectrum reaches from the computationally lightweight correlation of answers on a public dataset \cite{LW1_witt2021rewardbased} as a  proxy for contribution as opposed to the Shapley value, a measurement with strong theoretical properties but massive computational overhead \cite{LW9_FedCoin,MH9_He2021}. 

\textbf{Special Focus}: The aforementioned complexity as well as its novelty results in many open issues across a broad spectrum. Many works, therefore, focus on solving specific issues such as enhanced privacy \cite{LW2_Weng2021, LW8_Privacy_IoV,LW12_Rahmadika2021_unlinkable, MH8_Fadaeddini2019}, novel BC systems \cite{LW4_FL_IIoT,LW9_FedCoin}, bandwidth reduction \cite{MH10_Qu2021}, novel contribution measurements \cite{LW1_witt2021rewardbased, LW9_FedCoin, MH10_Qu2021} or game theory (e.g., \cite{MH2_Kang2019, LW11_RewardResponseGame}), as the major contribution which further complicates a holistic comparison of FLF.

\textbf{Thoroughness}: The analyzed papers also vary heavily in provided detail and thoroughness, ranging from first concepts, lacking details in terms of important specifications such as performance, specific function, operation and storage on BC, contribution measurement, robustness, experiments and privacy to theoretically detailed and experimentally tested solutions. None of the analyzed papers are production-ready.

\subsubsection{Standards for better comparability} For better reproducibility, implementability, and comparability we suggest considering and defining the following elements when designing a FLF.\newline

\noindent \textit{System model and architecture:}
\begin{itemize}
    \item Assumed application
    \item Type of FL (i.e., CD vs CS, horizontal vs vertical)
    \item Entities (including attackers)
    \item Setup (e.g., who manages a system, who deploys it)
    \item Role of BC within the FLF (e.g., what part does BC replace, what functions/operations)
    \item BC design (e.g., consensus algorithms, BCs, smart contracts)
    \item Non-BC design (e.g., off-chain storage, privacy protection, authentication)
    \item Procedures (e.g., flowcharts and diagrams)
    \item Theoretical analysis of incentive mechanisms 
    \item Specification of clients' contribution measurement
    \item Possible attacks (e.g., system security, data privacy)
\end{itemize}

\noindent \textit{Performance analysis:}
\begin{itemize}
    \item Quantitative performance analysis 
    \item Scalability analysis with respect to blockchain and contribution measurement 
\end{itemize}

\noindent \textit{Cost analysis:}
\begin{itemize}
    \item Overhead and cost analysis of BC infrastructure
    \item Overhead and cost analysis of the contribution measurement
    \item Performance-cost trade-off discussion
\end{itemize}

%% file: PRISMA_grafik_org.tex
\begin{tikzpicture}[thick, node distance=2em]
\def\x{width{"Records identified trough database search  Total records   1261"}}
\tikzset{
  rec/.style={
    rectangle,
    fill=accessDarkBlue,
    text=white,
    /utils/exec={\sffamily},
    minimum width=\x*1.2
  }
}
\tikzset{
  srec/.style={
    minimum width=\x*0.58
  }
}
\tikzset{
  svrec/.style={
    minimum width=\x*0.58,
    text width = \x*0.52,
    align=center
  }
}
\tikzset{
  excl/.style={
  fill=accessDarkBlue!50,
  }
}

\tikzset{
    barstyle/.style={
        fill=black!60,
        text=white,
        /utils/exec={\sffamily},
        align=center,
        inner ysep=6pt,
        rotate=90,
        rounded corners=1.2em,
        minimum height=2.5em
    }
}

\node[rec]
  (DB) {\begin{tabular}{c}Records identified through database search \textbf{(Total records = 422)} \\
  SpringerLink Database (n=167)\\
  ScienceDirect Database (n=95)\\
  ACM Digital Library (n=30) \\
  IEEE Xplore Digital Library (n=29) \\
  Emerald insight (n=29)  \\
  MDPI Database (n=27) \\
  Hindawi Database (n=20) \\
  Tayler \& Francis (n=18) \\
  SAGE Journals (n=5) \\
  Inderscience (n=2) \\
  Wiley Databease (n=0)  \\
  \end{tabular}};

\node[rec, 
    below= of DB.south]
  (dup) {\begin{tabular}{c} Records after duplicates removed \\ \textbf{(n = 422)} \end{tabular}};
  
\path let
\p1=(dup), \p2=(DB.north)
in
node[barstyle,left=of DB.north west,minimum width=\y2-\y1-0.1em] 
  (barID) {Searching};

\path (dup.north) -- (dup) coordinate[midway] (scex);

\node[rec, srec, below= of dup.south west,anchor=north west]
  (screen) {\begin{tabular}{c} Titles and abstracts screened \\ \textbf{(n = 422)} \end{tabular}}; 
  
\node[rec, srec, excl, below= of dup.south east,anchor=north east]
  (ex) {\begin{tabular}{c} Records excluded \\ \textbf{(n = 338)} \end{tabular}};

\path let
\p1=(screen.south), \p2=(dup)
in
node[barstyle,left=of dup,minimum width=\y2-\y1+1em] 
  (barScreen) {Inclusion};

\path (screen.south) -- (ex.south) coordinate[midway] (scex);  

\node[rec, excl, svrec, below=of ex, minimum height=10em]
  (FTout) {Full-text articles excluded with reasons \textbf{(Total = 45)}  \\ Other topic (n = 3) \\ No MD+FL+BC (n = 40)\\ Too general (n = 2) };

\node[rec, svrec, below=of screen, minimum height=10em]
  (FTin) { Full-text articles assessed for eligibility \\ \textbf{(n = 84)}};

\node[rec, srec, below=of FTin, minimum height=4.5em]
  (Elli) {\begin{tabular}{c} Eligible studies found \\ \textbf{(n = 39)} \end{tabular}};
\node[rec, svrec, below=of FTout, minimum height=4.5em]
  (Elliext) { Eligible studies found through references\\ \textbf{(n = 1)}};
  
\path let
\p1=(Elli), \p2=(FTin.north)
in
node[barstyle,left=of FTin.north west,minimum width=\y2-\y1-0.1em] 
  (barElli) {Eligibility Check};

\path (Elli.south) -- (Elliext.south) coordinate[midway] (FT);
\node[rec, below=of FT]
  (final) {\begin{tabular}{c} Studies included for quantitative and qualitative analysis \\ \textbf{(n = 40)} \end{tabular}};


\path let
\p1=(final.south), \p2=(Elli)
in
node[barstyle,left=of Elli,minimum width=\y2-\y1] 
  (barIncl) {Screening};
\path[->,draw=accessDarkBlue,line width=2pt]
    (DB) edge (dup)
    (dup) edge (screen)
    (screen) edge (ex)
    (screen) edge (FTin)
    (FTin) edge (FTout)
    (FTin) edge (Elli)
    (Elli) edge (final)
    (Elliext) edge (final);
\end{tikzpicture}

%

%% file: table_overview.tex
\rowcolors{3}{tableoddrow}{tableevenrow}
\begin{tabular}{D|DDQD|DDDDDD} 
    \rowcolor{tableheader} 
     & & & & & \multicolumn{6}{D}{Domains} \\
    \rowcolor{tableheader} 
    \multirow{-2}{*}{\cellcolor{tableheader} Ref.} & \multirow{-2}{*}{\cellcolor{tableheader} Application} & \multirow{-2}{*}{\cellcolor{tableheader} Setting} & \multirow{-2}{*}{\cellcolor{tableheader} Actors} & \multirow{-2}{*}{\cellcolor{tableheader} Setup} & SPoF & BC & FL & IM & CM & SP \\
    \hline
    \cite{LW1_witt2021rewardbased} & Generic & n.s. & Workers & \checkmark & \checkmark & & \checkmark & \checkmark & \checkmark &\\ 
    \cite{LW2_Weng2021} & Generic & CS & Workers & \checkmark & \checkmark & & & \checkmark & & \checkmark\\
    \cite{LW3_Kumar2020} & Generic & n.s. & Contributors, miners & \checkmark & \checkmark & & \checkmark & & & \checkmark\\
    \cite{LW4_FL_IIoT} & IoT & n.s. & Clients, central organization &  & \checkmark & & \checkmark & \checkmark & & \checkmark\\
    \cite{LW5_FL_HomeAppliances_IoT} & IoT & n.s. & Clients, miners & \checkmark & \checkmark & & & & & \checkmark\\
    \cite{LW6_Rahmadika2020} & Generic & n.s. & Aggregation servers, workers & \checkmark & & & & \checkmark & &\\
    \cite{LW7_Zhang_Reputation} & Generic & n.s. & Workers, task publishers, miners &  & & & & \checkmark & &\\
    \cite{LW8_Privacy_IoV} & IoV & n.s. & MEC, MBS, ADV &  & & & & \checkmark & & \checkmark\\
    \cite{LW9_FedCoin} & Generic & n.s. & Model requesters, FL servers, clients &  & & \checkmark & & \checkmark & &\\
    \cite{LW10_refiner} & Generic & CD & Administrator, requesters, workers, validators & \checkmark & & & & \checkmark & & \checkmark\\
    \cite{LW11_RewardResponseGame} & Generic & n.s. & Central server, workers &  & & &  & \checkmark & &\\
    \cite{LW12_Rahmadika2021_unlinkable} & Generic & CS & Workers, leaders, aggregation server &  & & & & \checkmark & & \checkmark\\
    \cite{LW13_TowardsReputationINFOCOMM} & Generic & n.s. & Edge devices, fog nodes, cloud & \checkmark & \checkmark & & & & &\\
    \cite{LW14_5G_Rahmadika2021} & Generic & CS & Users, edge devices, cloud &  & \checkmark & & & & & \checkmark\\
    \cite{MH1_Chai2021} & IoV & CD & Vehicles, RSUs, BSs &  & \checkmark & & & & &\\
    \cite{MH2_Kang2019} & Generic & CD & Task publishers, workers &  & & & & \checkmark & &\\
    \cite{Bao2019FLChain} & Generic & n.s. & Trainers, buyers, reporters, data processors & \checkmark & \checkmark & & & \checkmark & &\\
    \cite{MH4_Ma2021} & Generic & CS & Data owners & \checkmark & & & & \checkmark & &\\
    \cite{MH5_Lei2021} & IoT & CD & Local devices, BSs, MEC nodes & \checkmark & \checkmark & & & \checkmark & &\\
    \cite{MH6_Kansra2022} & IoV & CD & Vehicle edge nodes, BC nodes &  & \checkmark & & \checkmark & \checkmark & &\\
    \cite{MH7_Mugunthan2022} & Generic & n.s. & Trainer nodes & \checkmark & \checkmark & & & \checkmark & & \checkmark\\
    \cite{MH8_Fadaeddini2019} & Generic & n.s. & Model initiators, computing partners, validators &  & \checkmark & & & & &\\
    \cite{MH9_He2021} & Finance & CS & Follower candidates, leader nodes & \checkmark & \checkmark & \checkmark & & \checkmark & &\\
    \cite{MH10_Qu2021} & IoT & CD & Users, edge server, cloud server &  & & & & & &\\
    \cite{MH11_Desai2021} & Generic & CD & Worker nodes & \checkmark & \checkmark & & & & & \checkmark\\
    \cite{MH12_Zhu2021} & Generic & n.s. & Task publishers, parties, miners, smart contract & \checkmark & \checkmark & & & & & \checkmark\\
    \cite{MH13_Li2020} & Generic & n.s. & Requesters, workers, crowdsourcing platform &  & \checkmark & & & \checkmark & \checkmark & \checkmark\\
    \cite{MH14_Rathore2019} & IoT & n.s. & IoT devices, edge servers, central cloud server &  & \checkmark & & & & & \checkmark\\
    \cite{KT01_Toyoda2019BigData, KT02_Toyoda2020Access} & Generic & n.s. & Task requesters, workers & \checkmark & \checkmark & & & \checkmark & &\\
   \cite{KT03_Martinez2019CyberC} & Generic & n.s. & Requesters, workers &  & \checkmark & & & & \checkmark &\\
    \cite{KT04_Zou2021WCNC} & IoV & CD & RSUs (aggregators), MECs, vehicles (workers) &  & \checkmark &&& \checkmark & \checkmark &\\
    \cite{KT05_Hu2021IoTJ} & IoT & n.s. & Requesters, edge servers (workers), data collectors &  & \checkmark & & & \checkmark & & \checkmark\\
    \cite{KT06_Wang2021TNSE} & IoV & n.s. & UAVs (sensors, workers), requesters, MECs &  & \checkmark & \checkmark &&& \checkmark & \checkmark\\
    \cite{KT07_Ur_Rehman2020INFOCOMW} & Generic & n.s & Requesters, data-arbitrators, workers &  & \checkmark & & & & &\\
    \cite{KT08_Qu2021TPDS} & Generic & n.s. & Requesters, miners &  & \checkmark & \checkmark & & \checkmark & \checkmark &\\
    \cite{KT09_Zhang2021IJHPSA} & Generic & n.s. & Miners, workers &  & & \checkmark & & & &\\
    \cite{KT10_Gao2021ICPP} & Generic & n.s. & Requesters, workers, aggregation servers &  & \checkmark & & & \checkmark & \checkmark & \checkmark\\
    \cite{KT11_Xuan2021SCN} & Generic & n.s. & Administrator, requesters, workers, miners &  & \checkmark & & & \checkmark & &\\
    \cite{KT12_Liu2021Sensors} & Generic & CS, CD & Workers, miners &  & \checkmark & & \checkmark & & &\\
\end{tabular}

%% file: table_blockchain.tex
\rowcolors{2}{tableoddrow}{tableevenrow}
\begin{tabular}{D|DDDDQQ|QDD|D}
    \rowcolor{tableheader}
    & \multicolumn{4}{D}{Operations} & & & \multicolumn{2}{D}{On-chain} & & \\
    \rowcolor{tableheader} 
    \multirow{-2}{*}{\cellcolor{tableheader} Ref.} & 
    Agg. & Cor. & Pay. & Str. &
    \multirow{-2}{*}{\cellcolor{tableheader} BC} &
    \multirow{-2}{*}{\cellcolor{tableheader} Consensus} &
    Items & Eval. & 
    \multirow{-2}{*}{\cellcolor{tableheader} Off-chain} &
    \multirow{-2}{*}{\cellcolor{tableheader} Scalability}
    \\
    \hline
   \cite{LW1_witt2021rewardbased} & \checkmark & \checkmark & \checkmark & & Agnostic & n.a. & 1-bit results of all participants & \checkmark & & \checkmark \\
   \cite{LW2_Weng2021} & \checkmark & & \checkmark & & Corda V3.0 & Algorand & Gradients & & & \\
   \cite{LW3_Kumar2020} & & & \checkmark & & Ethereum & PoW & n.s. & & \checkmark & \\
   \cite{LW4_FL_IIoT} & & & \checkmark & & Ethereum & PoW & Contribution, Merkle tree & & & \\
   \cite{LW5_FL_HomeAppliances_IoT} & & \checkmark & \checkmark & & n.s. & n.s. & Models & & \checkmark & \\
   \cite{LW6_Rahmadika2020} & & & \checkmark & & Ethereum & PoW & n.s. & \checkmark & & \\
   \cite{LW7_Zhang_Reputation} & & & & \checkmark & TrustRE & PoR & Reputation scores & & & \\
   \cite{LW8_Privacy_IoV} & & & & \checkmark & Custom & PoW & Model updates & & & \\
   \cite{LW9_FedCoin} & & & \checkmark & & Custom & PoSap & SV values, tasks & & & \\
   \cite{LW10_refiner} & & \checkmark & \checkmark & \checkmark & Ethereum & PoW & Tasks & & \checkmark & \\
   \cite{LW11_RewardResponseGame} & \checkmark & & \checkmark & \checkmark & n.s. & PoW & Model updates, metadata & & & \\
   \cite{LW12_Rahmadika2021_unlinkable} & & & \checkmark & & Ethereum, HF & PoW & PoT records & \checkmark & & \\ 
   \cite{LW13_TowardsReputationINFOCOMM} & & \checkmark & & \checkmark & n.s. & n.s. & Reputation scores & & \checkmark & \\ 
   \cite{LW14_5G_Rahmadika2021} & & \checkmark & \checkmark & & Ethereum & PoW & Users' addresses & & & \\
   \cite{MH1_Chai2021} & & & \checkmark & \checkmark & n.s. & PoK & Local models, loss, signatures & & & \checkmark \\
   \cite{MH2_Kang2019} & & & & \checkmark & Corda V4.0 & PBFT & Reputation scores & & \checkmark & \\
   \cite{Bao2019FLChain} & & \checkmark & \checkmark & \checkmark & n.s. & Custom & Client info, model parameters & & & \checkmark \\
   \cite{MH4_Ma2021} & \checkmark & & & & n.s. & n.s. & Masked gradients, global models & & & \\
   \cite{MH5_Lei2021} & \checkmark & & \checkmark & \checkmark & HF & Raft & Local updates & & & \checkmark \\
   \cite{MH6_Kansra2022} & & \checkmark & \checkmark & & n.s. & n.s. & & & & \\
   \cite{MH7_Mugunthan2022} & & \checkmark & \checkmark & & Ethereum & PoW & IPFS CIDs of models & & \checkmark & \\
   \cite{MH8_Fadaeddini2019} & & \checkmark & \checkmark & & Stellar & n.s. & IPFS CIDs of models & & \checkmark & \\
   \cite{MH9_He2021} & \checkmark & \checkmark & & \checkmark & Custom & Raft & Local and global models, loss & & & \\
   \cite{MH10_Qu2021} & & & & \checkmark & n.s. & Custom & Local and global models & & & \\
   \cite{MH11_Desai2021} & \checkmark & \checkmark & & & HF & Raft & Parameters & & & \checkmark \\
   \cite{MH11_Desai2021} & & & \checkmark & \checkmark & Ethereum & PoW & Hashes of parameters & & & \checkmark \\
   \cite{MH12_Zhu2021} & & \checkmark & \checkmark & \checkmark & Ethereum & PoW & Aggregated models & & & \\
   \cite{MH13_Li2020} & & \checkmark & & \checkmark & Ethereum & PoW & Encrypted models & \checkmark & & \checkmark \\
   \cite{MH14_Rathore2019} & & \checkmark & & & Ethereum & PoW & Aggregated local updates & & & \\
   \cite{KT01_Toyoda2019BigData, KT02_Toyoda2020Access} & \checkmark & \checkmark & \checkmark & \checkmark & Agnostic & n.a. & Tasks, voting results, model updates & & \checkmark & \\
   \cite{KT03_Martinez2019CyberC} & & & \checkmark & \checkmark & EOS, HF & n.s. & Hashes of model updates, data size & & \checkmark & \\
   \cite{KT04_Zou2021WCNC} & & \checkmark & \checkmark & \checkmark & n.s. & n.s. & Models & & & \\
   \cite{KT05_Hu2021IoTJ} & & & & \checkmark & Agnostic & PBFT & Model updates & & & \\
   \cite{KT06_Wang2021TNSE} & & & & \checkmark & n.s. & PoW & Tasks, model updates, aggregated models & & & \\
   \cite{KT07_Ur_Rehman2020INFOCOMW} & & & & \checkmark & Ethereum & PoW & Reputation scores & & \checkmark & \\
   \cite{KT08_Qu2021TPDS} & & & \checkmark & \checkmark & Custom & PoFL & Model updates & & & \\
   \cite{KT09_Zhang2021IJHPSA} & & & \checkmark & \checkmark & Custom & PoMC & Model updates & & & \\
   \cite{KT10_Gao2021ICPP} & & & & \checkmark & n.s. & n.s. & Signatures, reputation scores, contributions & & & \\
   \cite{KT11_Xuan2021SCN} & & \checkmark & \checkmark & \checkmark & Agnostic & n.s. & Tasks, voting results, model updates & & & \\
   \cite{KT12_Liu2021Sensors} & \checkmark & & \checkmark & \checkmark & Custom & PoW & Model updates, computation time & & & \\
\end{tabular}

%% file: table_incentive_mechanism.tex
\rowcolors{2}{tableoddrow}{tableevenrow}
\begin{tabular}{D|QQQ|QDDDQ}
    \rowcolor{tableheader} 
     & & & & \multicolumn{5}{D}{Contribution} \\
    \rowcolor{tableheader} 
    \multirow{-2}{*}{\cellcolor{tableheader} Ref.} & 
    \multirow{-2}{*}{\cellcolor{tableheader} Sim.} & 
    \multirow{-2}{*}{\cellcolor{tableheader} Theoretical analysis} & 
    \multirow{-2}{*}{\cellcolor{tableheader} Costs} & 
    Metrics & Abs. & Rel. & Rep. & Validator \\
    \hline
    \cite{LW1_witt2021rewardbased} & \checkmark & \checkmark & Generic & Correlation of predictions (Peer truth serum) & & \checkmark & & Smart contract \\
    \cite{LW2_Weng2021} &  & \checkmark & n.s. & n.s. & & & & Miners \\
    \cite{LW3_Kumar2020} &  &  & n.a. & Accuracy (validation scores) & \checkmark & & & Miners \\
    \cite{LW4_FL_IIoT} & \checkmark &  & n.a. & Accuracy, data size & \checkmark & & & Agg. server \\
    \cite{LW5_FL_HomeAppliances_IoT} & \checkmark &  & n.a. & Euclidean distance of model updates & & \checkmark & & Miners \\
    \cite{LW6_Rahmadika2020} & \checkmark &  & n.a. & Data size & \checkmark & & & Smart contract \\
    \cite{LW7_Zhang_Reputation} & \checkmark &  &  n.a. & Accuracy, energy consumption, data size & & \checkmark & \checkmark & Task requesters \\
    \cite{LW8_Privacy_IoV} &  &  & n.a.  & Accuracy (loss) & \checkmark & & & MEC servers \\
    \cite{LW9_FedCoin} & \checkmark &  & n.a. & Accuracy (loss) (Shapley values) & & \checkmark & & Miners \\
    \cite{LW10_refiner} &  &  & n.a. & Accuracy (loss, marginal) (rank) & & \checkmark & & Validators \\
    \cite{LW11_RewardResponseGame} & \checkmark & \checkmark(Stackelberg game) & Computation & Accuracy, data size & \checkmark & & & Agg. servers \\
    \cite{LW12_Rahmadika2021_unlinkable} &  &  & n.a. & n.s. & & & & n.s. \\
    \cite{LW13_TowardsReputationINFOCOMM} &  &  & n.a. & n.s. & & & & n.s. \\
    \cite{LW14_5G_Rahmadika2021} &  &  & n.a. & Data size & \checkmark & & & Task requesters \\
    \cite{MH1_Chai2021} & \checkmark & \checkmark(Stackelberg game) & Computation & Accuracy (loss) & \checkmark & & & Agg. servers \\
    \cite{MH2_Kang2019} & \checkmark & \checkmark(Contract theory) & Energy & RONI \cite{Biscotti2018}, FoolsGold \cite{Fung2018FoolsGold} & & & \checkmark & Task requesters \\
    \cite{Bao2019FLChain} &  & \checkmark & Generic & n.s. & & & & Workers \\
    \cite{MH4_Ma2021} & \checkmark &  & n.a. & Generic (Shapley values) & & \checkmark & & Smart contract \\
    \cite{MH5_Lei2021} &  &  & n.a. & n.s. & & & & n.s. \\
    \cite{MH6_Kansra2022} &  &  & n.a. & Accuracy (generic, marginal) & \checkmark & & & n.s. \\
    \cite{MH7_Mugunthan2022} & \checkmark &  & n.a. & Accuracy (generic) & & \checkmark & & Workers \\
    \cite{MH8_Fadaeddini2019} &  &  & n.a. & n.s. & & & & Validators \\
    \cite{MH9_He2021} & \checkmark &  & n.a. & Similarity of model updates (Shapley values) & & \checkmark & & n.s. \\
    \cite{MH10_Qu2021} & \checkmark &  & n.a. & Communication delay, energy consumption & \checkmark & & & Agg. servers \\
    \cite{MH11_Desai2021} &  &  & n.a. & Speed of model submission & \checkmark & & & n.s. \\
    \cite{MH12_Zhu2021} &  &  & n.a. & n.s. & & & & n.s. \\
    \cite{MH13_Li2020} &  &  & n.a. & Accuracy, data size & \checkmark & & \checkmark & Task requesters \\
    \cite{MH14_Rathore2019} &  &  & n.a. & Data size & \checkmark & & & Agg. servers \\
    \cite{KT01_Toyoda2019BigData, KT02_Toyoda2020Access} &  & \checkmark(Contest theory) & Computation & Accuracy (generic) (rank by voting) & & \checkmark & & Workers \\
    \cite{KT03_Martinez2019CyberC} &  &  & n.a. & Data size & \checkmark & & & Workers \\
    \cite{KT04_Zou2021WCNC} & \checkmark & \checkmark & Data, communication & Accuracy (loss) & \checkmark & & & Agg. servers \\
    \cite{KT05_Hu2021IoTJ} & \checkmark & \checkmark & Sensing, privacy & n.s. & & & & n.s. \\
    \cite{KT06_Wang2021TNSE} & \checkmark & \checkmark(Reinforcement learning) & Sensing, privacy, energy & Data size, sensing capacity & \checkmark & & & Agg. servers \\
    \cite{KT07_Ur_Rehman2020INFOCOMW} &  &  & n.a. & n.s. & & & & n.s. \\
    \cite{KT08_Qu2021TPDS} & \checkmark & \checkmark & Privacy & Accuracy & \checkmark & & & BC nodes \\
    \cite{KT09_Zhang2021IJHPSA} &  &  & n.a. & n.s. & & & & n.s. \\
    \cite{KT10_Gao2021ICPP} & \checkmark & \checkmark & n.s. & Accuracy (loss) & \checkmark & & \checkmark & Workers \\
    \cite{KT11_Xuan2021SCN} &  &  & n.a. & Accuracy (generic) (rank by voting) & & \checkmark & & Workers \\
    \cite{KT12_Liu2021Sensors} &  &  & n.a. & Computation time & \checkmark & & & Miners \\
\end{tabular}

%% file: table_experiments.tex
\rowcolors{2}{tableoddrow}{tableevenrow}
\begin{tabular}{D|DDQEQ|QDDDDDDDD}
    \rowcolor{tableheader} 
     & \multicolumn{2}{D}{Tasks} & & & & & & \multicolumn{4}{D}{Adversaries} & & \\
    \rowcolor{tableheader}
    \multirow{-2}{*}{\cellcolor{tableheader} Ref.} & 
    Clf. & Rgr. &
    \multirow{-2}{*}{\cellcolor{tableheader} Datasets} & 
    \multirow{-2}{*}{\cellcolor{tableheader} \#Clients} & 
    \multirow{-2}{*}{\cellcolor{tableheader} Algorithms} & 
    \multirow{-2}{*}{\cellcolor{tableheader} Privacy} & 
    \multirow{-2}{*}{\cellcolor{tableheader} Non-iid} & 
    BT & RP & RT & SP & 
    \multirow{-2}{*}{\cellcolor{tableheader} Imp.} &
    \multirow{-2}{*}{\cellcolor{tableheader} Per.} \\
    \hline
    \cite{LW1_witt2021rewardbased} & \checkmark & & EMNIST & 10 & FD &  & \checkmark & & \checkmark & & \checkmark &  & \checkmark \\
    \cite{LW2_Weng2021} & \checkmark & & MNIST & 4-10 & FedAvg & HE (Paillier) &  & & & & & \checkmark & \checkmark \\
    \cite{LW3_Kumar2020} & \checkmark & & MNIST & 5 & FedAvg, EWC & DP, HE  & \checkmark & & & & &  & \checkmark \\
    \cite{LW4_FL_IIoT} & \checkmark & & Original & 4 & FedAvg, CDW & n.s. & \checkmark & & & & & \checkmark & \checkmark \\
    \cite{LW5_FL_HomeAppliances_IoT} & \checkmark & & MNIST & 10 & FedAvg & DP &  & & \checkmark & & &  & \checkmark \\
    \cite{LW6_Rahmadika2020} & \checkmark & & MNIST & 25 & FedAvg & n.s. &  & & & & & \checkmark & \checkmark \\
    \cite{LW7_Zhang_Reputation} & \checkmark & & MNIST, CIFAR10 & n.s. & n.s. &  &  & & & & &  & \checkmark \\
    \cite{LW8_Privacy_IoV}  & & \checkmark & Real-time AD video & 1 & n.s. & HE (DGHV), ZKP &  & & & & &  & \checkmark \\
    \cite{LW9_FedCoin} & \checkmark & & MNIST & n.s. & FedAvg & SA &  & & & & & \checkmark &  \\ 
    \cite{LW10_refiner} & \checkmark & & MNIST & 5 & FedAvg, FedProx & Sym. cryptography &  & & \checkmark & & & \checkmark &  \\
    \cite{LW11_RewardResponseGame} & \checkmark & & Reddit, Celeba & 5-75 & n.s. &  &  & & & & &  & \checkmark \\
    \cite{LW12_Rahmadika2021_unlinkable} & \checkmark & & MNIST & 100 & Original & Ring sig., HE (RSA), Rabin, ECC &  & & & & & \checkmark & \checkmark \\
    \cite{LW13_TowardsReputationINFOCOMM} & & & n.a. & n.a. & n.a. &  &  & & & & &  &  \\ 
    \cite{LW14_5G_Rahmadika2021} & \checkmark & & Mathworks handwritten & 10 & FedAvg & Pairing-based cryptography, ECC &  & & & & & \checkmark & \checkmark \\
    \cite{MH1_Chai2021} & \checkmark & & MNIST, CIFAR10 & 6 & Original & Asym. cryptography, signatures &  & & \checkmark & & &  & \checkmark \\
    \cite{MH2_Kang2019} & \checkmark & & MNIST & 100 & FedAvg &  & \checkmark  & & \checkmark & \checkmark & & \checkmark & \checkmark \\
    \cite{Bao2019FLChain} & \checkmark & & n.s. & 10 & n.s. & n.s. &  & & & & & \checkmark & \checkmark \\
    \cite{MH4_Ma2021} & \checkmark & & ORHD & 9 & FedAvg & SA  & \checkmark & & & & &  &  \\
    \cite{MH5_Lei2021} & \checkmark & & MNIST & 30 & n.s. &  &  & & & & &  & \checkmark \\
    \cite{MH6_Kansra2022} & & & n.a. & n.a. & n.a. &  &  & & & & &  &  \\
    \cite{MH7_Mugunthan2022} & \checkmark & & Adult census, KDD & 50 & Custom FedAvg & DP & \checkmark & & \checkmark & \checkmark & \checkmark &  & \checkmark \\
    \cite{MH8_Fadaeddini2019} & & & n.a. & n.a. & n.a. &  &  & & & & &  &  \\
    \cite{MH9_He2021} & & & n.a. & n.a. & n.a. &  &  & & & & &  &  \\
    \cite{MH10_Qu2021} & \checkmark & & Original  & 3, 4 & FedAvg & n.s. &  & & \checkmark & & &  & \checkmark\\
    \cite{MH11_Desai2021} & \checkmark & & CIFAR10 & 100 & FedAvg, signSGD & n.s. & \checkmark & & & & \checkmark & \checkmark &  \\
    \cite{MH12_Zhu2021} & \checkmark & & FMNIST & 900 & FedAvg & HE (Paillier) & \checkmark & & \checkmark & & & \checkmark & \checkmark \\
    \cite{MH13_Li2020} & \checkmark & & BCWD, HDD & 3 & FedAvg & DP &  & & & & & \checkmark & \checkmark \\
    \cite{MH14_Rathore2019} & \checkmark & \checkmark & PASCAL VOC 2012 & 5-10 & FedAvg & HE &  & & & & & \checkmark & \checkmark \\
    \cite{KT01_Toyoda2019BigData, KT02_Toyoda2020Access} & & & n.a. & n.a. & n.a. &  &  & & & & &  &  \\
    \cite{KT03_Martinez2019CyberC} & \checkmark & & MNIST & 10 & FedAvg &  & \checkmark & & & & & \checkmark & \checkmark \\
    \cite{KT04_Zou2021WCNC} & \checkmark & & MNIST & 50 & n.s. &  &  & & & & &  & \checkmark \\
    \cite{KT05_Hu2021IoTJ} & \checkmark & & FEMNIST & 35, 105, 175 & n.s. &  & \checkmark & & & & &  & \checkmark \\
    \cite{KT06_Wang2021TNSE} & \checkmark & & MNIST & n.s.\footnote{Only the density of UAVs is given} & n.s. &  &  & & & & &  &  \\
    \cite{KT07_Ur_Rehman2020INFOCOMW} & & & n.a. & n.a. & n.a. &  &  & & & & &  &  \\
    \cite{KT08_Qu2021TPDS} & \checkmark & & CIFAR-10 & 20, 50, 100 & n.s. & HE, 2PC &  & & & & &  & \checkmark \\
    \cite{KT09_Zhang2021IJHPSA} & \checkmark & & ImageNet & 20 & n.s. &  &  & \checkmark & \checkmark & & & \checkmark & \checkmark \\
    \cite{KT10_Gao2021ICPP} & \checkmark & & MNIST, CIFAR-10 & 10 & FedAvg &  &  & & \checkmark & & \checkmark &  & \checkmark \\
    \cite{KT11_Xuan2021SCN}& \checkmark & & MNIST & 50 & FedAvg &  &  & & & & &  & \checkmark \\
    \cite{KT12_Liu2021Sensors} & \checkmark & & MNIST, CIFAR-10 & 2-6 & FedAvg &  & \checkmark  & & & & &  & \checkmark \\
\end{tabular}

%% file: 5_future_research_directions.tex
\section{Future Research Directions}
\label{sec:future_research_directions}
The multitude of possible applications of FLF come with different requirements in terms of accuracy, latency, cost, and privacy. To account for this, we classify future research into two main directions, namely (i) increase in framework performance and (ii) expansion of framework functionalities.


\subsection{Performance}
Most state-of-the-art publications only consider BC and incentive mechanisms on a conceptual or theoretical level, however, they lack a performance analysis. Yet, low operational costs and latency, as required by real-time applications, such as autonomous driving, demand high-performance systems. To develop such frameworks, we have identified four performance bottlenecks as future research directions: (i) framework scalability, (ii) communications and network, (iii) framework implementation, and (iv) framework evaluation and comparison.

\subsubsection{Framework scalability}
One of the major factors for the applicability of a FLF is its ability to scale beyond small groups toward mass adoption. Out of the \finalresults papers, only six mentioned and considered scalability within the design of their respective FLF. In particular, our reviews show that the integration of distributed ledger technology frequently leads to scalability problems. In FLFs, BC technology becomes a scalability bottleneck if
\begin{enumerate}
    \item it is part of the operating core infrastructure of the FLF (e.g., \cite{LW1_witt2021rewardbased}) and not only a complementary outer layer technology (e.g., \cite{LW7_Zhang_Reputation})
    \item heavy operations such as aggregation or reward calculation are performed on-chain \cite{LW9_FedCoin} 
    \item a large amount of information is stored on the BC such as model updates 
    \item the BC framework is public and used outside the realm of the FLF 
    \item the consensus mechanism is resource-intense (e.g., PoW).
\end{enumerate}
There are multiple promising future strategies to improve the scalability of the framework. First, FLF-specific BC systems have been proposed that replace the computational overhead of the PoW-based systems with computational heavy tasks in FLF such as model parameter verification~\cite{KT08_Qu2021TPDS}, reputation verification~\cite{LW7_Zhang_Reputation}, or contribution measurement calculations~\cite{LW9_FedCoin}. Secondly, Zhang~\etal~\cite{KT09_Zhang2021IJHPSA} have investigated the use of efficient AI hardware to increase BC scalability. Wang~\etal explored the domain of resource optimization in BC-based FLFs to further improve the scalability \cite{ResourceOptimization}. Moreover, Weng~\etal~\cite{LW2_Weng2021} aim to improve scalability by enhancing the privacy procedures for the FLF processes. Another promising research direction is the application of Zero-Knowledge Succinct Non-Interactive Argument of Knowledge (ZK-SNARKs)~\cite{Pinto2020zkSNARKs} in the FLF context. ZK-SNARKs is a promising cryptographic technology that allows a \textit{prover} to prove to a \textit{verifier} that computation has been executed without revealing the program itself. This verification is faster than actually computing the original code and can be implemented easily on the smart contract. Hence, this will improve the scalability of BC-enabled FLF dramatically. However, due to its generality, which processes leverage ZK-SNRAKs is an open question. Finally, the performance of the BC itself can be improved, e.g., by increasing the number of transactions per second.

 
\subsubsection{Communication and network}
Another major remaining challenge is the communication bottleneck. Decentralized wireless gadgets, as employed in decentralized FL, operate on lower communication rates than traditional intra- or inter-datacenter links. This leads to a trade-off between accuracy and communication cost. Although there exists first theoretical research on the nature of this trade-off, its findings have not yet been incorporated in the proposed FL frameworks~\cite{AdvancesAndOpenProblemsInFederatedLearning}. In terms of communication rates, new developments are also expected once 6G technology is introduced, which is predicted to be mutually empowering with FL~\cite{Liu20206G}. 

Furthermore, researchers face the communication-related problem of scheduling and resource allocation under dynamic channel condition and heterogeneous computing capacity of devices in IoT~\cite{yang2021privacy}. For instance, Yang~\etal~propose a device selection strategy in UAV to keep the low-quality devices from affecting the learning efficiency and accuracy~\cite{yang2021privacy}.

Another challenge is related to key collisions during update communication: To avoid throughput issues, data is typically uploaded iteratively in multiple smaller batches, causing latency and collision effects to become more dominant. For instance, Desai~\etal~\cite{MH11_Desai2021} point out that Hyperledger cannot deal with the Multiversion Concurrency Control (MVCC) of its underlying database so many transactions fail and need to be repeated. Accordingly, future research should be directed to the three compression objectives identified by Kairouz~\etal~\cite{AdvancesAndOpenProblemsInFederatedLearning}: gradient compression (client-to-aggregator communication), model compression (aggregator-to-client communication), and local computation reduction. In consequence, security and privacy mechanisms need to be adapted to operate on the compressed data (Section~\ref{sec:results_experiments_privacy}). Starting points for this upcoming research include sparsification and quantization approaches~\cite{Li.2020}, 1-bit compression~\cite{LW1_witt2021rewardbased}, or the parallelization on multiple contracts~\cite{MH1_Chai2021, MH5_Lei2021, MH11_Desai2021}. For the latter,  Desai~\etal~\cite{MH11_Desai2021} analyze the trade-off between communication speed and the number of employed parallelized contracts.

In general, future framework proposals should consider communication cost and time in their simulative methods, e.g., building up on Kang~\etal~\cite{MH2_Kang2019}.

\subsubsection{Framework implementation}
Most of the papers we reviewed are focused on the algorithm side. However, in order to go beyond theory towards real production-ready deployments, implementation details have to be taken into consideration. For instance, incurred deployment and maintenance costs of unproven novel BC systems are often ignored. Introducing a new, custom-made, and highly complex infrastructure introduces security risks and it requires a large team of experts to run and maintain such a system in practice. 
Therefore, software/hardware co-design is another vital topic in FL (e.g.,~\cite{khan2020federated, khan2021federated, guo2017software, KT09_Zhang2021IJHPSA}). For instance, Wang~\etal point out that cipher-text operation and encryption parts are major bottlenecks on the FL and proposed a novel Field Programmable Gate Array (FPGA) design for it~\cite{wang2022pipefl}. We believe that there are potential research topics in the software/hardware co-design for FL. Interested readers may refer to the survey papers of Khan~\etal~\cite{khan2020federated, khan2021federated}.

\subsubsection{Framework evaluation and comparison}
While many papers have conducted performance evaluation, few showed a comparison with other FLFs. This hinders the scientific advancement towards high-performing frameworks as the different design choices of the papers remain uncompared. Furthermore, the frameworks have often not been evaluated in realistic scenarios: \one relatively well-known benchmark datasets such as MNIST and CIFAR-10 are chosen (29 out of 34 papers that conducted experiments on classification) and \two the non-IID setting is only applied in 11 out of 34 papers. Furthermore, inconsistencies between the targeted FL setting (i.e., CS, CD) and the number of clients in the experiments are observed. In particular, FLFs that assume CD should simulate a large number of clients, however, only Kang~\etal~\cite{MH2_Kang2019} and Desai~\etal\cite{MH11_Desai2021} conducted experiments with 100 participants or more (\tablename~\ref{table:experiments}).

To better evaluate FLFs, we suggest using common datasets dedicated to FL (e.g., LEAF~\cite{Caldas2018LEAF}) as well as simulating different levels of non-IID data among clients (e.g., Dirichlet distribution \cite{LW1_witt2021rewardbased}). We also suggest deploying a FLF on the clusters of inexpensive computers such as Raspberry Pi~\cite{wang2021raspberrypi} to realistically simulate large-scale FL scenarios under the CD assumption.

Moreover, it is difficult to simulate the effect of decentralization and incentivization (e.g., Shapley value and game-theoretic mechanisms) in a comparable way since each paper uses different assumptions. Therefore, to fairly compare FLFs, holistic experiments should be designed, where the effects of decentralization and incentivization are captured by metrics such as overall accuracy, cost, or latency.

\subsection{Functionalities}

Future research should also focus on integrating further functionalities into the FLFs. Firstly, most of the proposed FL systems are limited to supervised classification, however, other types of ML problems should be considered as well. Secondly, lightweight privacy-preserving techniques are necessary for some applications that use sensitive information (e.g., medical logs and personal financial information). Thirdly, a fair, non-manipulable, and lightweight mechanism for contribution measurement has yet to be developed.

\subsubsection{Beyond supervised FL and federated averaging}
To expand the applicability of FLFs, machine learning tasks beyond supervised learning should be enabled, such as anomaly detection, reinforcement learning, natural language processing, user behavior analysis, and unsupervised learning tasks (e.g., \cite{Lin2021FedNLP, Servetnyk2020Unsupervised}). This will require new or adapted model aggregation algorithms and a new contribution measurement to integrate such tasks with IM and BC.

So far, \texttt{FedAvg} requires the same neural network architecture on all devices to participate. This may lead to issues in real-world environments where clients might have different hardware and bandwidth capabilities. Federated Knowledge Distillation~\cite{LW1_witt2021rewardbased} is an interesting novel FL approach in this context, allowing for a flexible neural network architecture and a dramatic reduction in bandwidth~\cite{sattler2020communicationefficient}. However, Federated Knowledge Distillation requires a public dataset to distill the knowledge. 

Traditional deep learning algorithms such as DNNs and Convolutional Neural Networks (CNNs), are generally power-hungry, which is problematic in IoT environment. To address this challenge, biological neurons-inspired DNNs called Spiking Neural Networks (SNNs) have been actively studied for edge AI (e.g., \cite{schranghamer2020graphene, yang2022lead}). SNNs will enable edge devices to exploit brain-like biophysiological structure to collaboratively train a global model while helping preserve privacy. For instance, Lead Federated Neuromorphic Learning (LFNL) is a method to enable SNNs in a federated manner~\cite{yang2022lead}. Furthermore, a leader election scheme is proposed to elect one device with high capability (e.g., computation and communication capabilities) as a leader to manage model aggregation, eliminating a fixed central coordinator and avoiding model poisoning attacks.

\subsubsection{Towards lightweight privacy-preserving FL}
Despite FL being a data privacy-preserving technology by design, research has shown that certain characteristics of the underlying training data sets can be inferred from the global model and that additional privacy-preserving measures are recommended. Our review shows that two classes of security concerns are targeted by the publications, namely \one leakage of data set characteristics and \two disclosure of participant identities. Although a substantial number of papers (20 out of 40 publications) address one of these concerns, only a single paper addresses both~\cite{LW8_Privacy_IoV}. Moreover, preventing data set leakage through DP or MPC inflicts trade-offs. Specifically, DP comes with a trade-off between data security and model accuracy, while MPC comes with a trade-off between data security and computation complexity, and it might thus not be applicable with a large number of participants~\cite{MH4_Ma2021}. It is worth noting that the model accuracy of DP cannot be inherently improved due to intentionally added noise. Hence, it would be important to explore lightweight MPC algorithms \cite{MH_SECURE_AGG} to accommodate a large number of clients for privacy-preserving FL. 




\subsubsection{Towards fair, non-manipulable, and lightweight contribution measurement}
Although multiple approaches for contribution measurements have been explored in the literature (Section~\ref{sec:results_incentive_mechanism}), a fair, non-manipulable, and lightweight
mechanism has yet to be
developed as the following overview shows. 

Firstly, a contribution can be measured based on the clients' honest reports of the amount of data, local accuracy, or local loss. Yet reward systems based on such simplified assumptions may not be applicable in any real-world scenario as the dominant strategy for an individual-rational agent is dishonest behavior (e.g., reporting the best possible outcome without costly model training). Recent technologies such as TEE and ZK-SNARKs are promising for trusted computation on mobile, edge, and IoT devices~\cite{oliveira2022utango}. However, how to leverage them to achieve honest reports without incurring additional costs (e.g., computational costs) is an open question.

Secondly, relative contribution measurement based on the client's reputation or majority voting is an interesting research avenue, promising to relax heavy verification and control mechanics for high-reputation clients. However, how to quantify the reputation fairly and robustly remains open research. Similarly, the majority voting methods may not reflect actual contribution due to its nature.

Thirdly, absolute or direct contribution measurement refers to assessing each client's model update on a public dataset. However, this approach \one requires a trusted central authority performing tests, and \two limits scalability due to the computational overhead. For instance, the Shapley value is a common method for measuring an agent's contribution, but still comes at the cost of heavy computational overhead even when optimized (e.g., ~\cite{shapleyeval, wang2020principled}).

Lately, correlation-based reward mechanisms, such as Correlated Agreement (CA) \cite{LIU_CA, CA_HONGTAO} and peer-truth serum \cite{LW1_witt2021rewardbased}, have been proposed as promising approaches for contribution measurements for FL. Without having access to the ground truth, the reward is calculated based on the correlation of the reported signals of peers. This implicit approach does not require an explicit contribution measurement and therefore avoids computational overhead.

In addition, when theoretically developing and analyzing incentive mechanisms, as performed by 12 out of 40 papers, more sophisticated assumptions concerning \one information availability \two uniformity in utility functions or \three individual rationality should be made to guarantee the robustness of the mechanisms in a real-world scenario. Specifically, as clients are humans, they may not follow their optimal strategies derived from the analysis. For instance, not all clients would take the cost of energy consumption into account when determining their strategies. We suggest taking humans' behavioral bias (e.g., prospect theory \cite{Kahneman1979, Tversky1992}) as well as non-quantifiable measures (e.g., the utility of privacy) into the theoretical analysis of incentive mechanisms. 

%% file: 6_conclusion.tex
\section{Conclusion and Outlook}
\label{sec:conclusion}
FL is a promising new AI paradigm focused on confidential and parallel model training on the edge. To apply FL beyond small groups of entrusted entities, a decentralization of power, as well as compensation for participating clients, has to be incorporated into the FLF. This work traversed and analyzed 12 leading scientific databases for incentivized and decentralized FLFs based on the PRISMA methodology, ensuring transparency and reproducibility. We found \rawresults papers and studied \finalresults works in-depth after three filtering rounds. To ensure correctness, the results were verified by the respective authors. We overcame the challenge of heterogeneity of FLFs in terms of use cases, applied focus, design choice, and thoroughness by offering a comprehensive and holistic comparison framework. By exposing the limitations of existing FLFs and providing directions for future research, this work aims to enhance the proliferation of incentivized and decentralized FL in practice.
